\documentclass[journal]{IEEEtran}
\usepackage{amsmath,amsfonts}
\usepackage{algorithmic}
\usepackage{algorithm}
\usepackage{array}
\usepackage[caption=false,font=normalsize,labelfont=sf,textfont=sf]{subfig}
\usepackage{textcomp}
\usepackage{stfloats}
\usepackage{url}
\usepackage{verbatim}
\usepackage{graphicx}
\usepackage{cite}
\usepackage{times}
\usepackage{epsfig}
\usepackage{graphicx}
\usepackage{amsmath}
\usepackage{amssymb}
\usepackage{overpic}
\usepackage{cite}
\usepackage{enumitem}

\usepackage{multirow}
\usepackage{booktabs}
\usepackage{color}
\usepackage{colortbl}
\definecolor{mygray}{gray}{.9}
\definecolor{mypink}{rgb}{.99,.91,.95}
\definecolor{mycyan}{cmyk}{.3,0,0,0}

\begin{document}

\title{Robust Single Object Tracking in LiDAR Point Clouds under Adverse Weather Conditions}

\author{Xiantong~Zhao,
        Xiuping~Liu,
        Shengjing~Tian*,
        Yinan~Han
        % <-this % stops a space
\thanks{Manuscript received --; revised --}
\thanks{Xiantong Zhao and Xiuping Liu are with the School of Mathematical Sciences, Dalian University of Technology, China (e-mail: xtongz.dut@gmail.com, xpliu@dlut.edu.cn).}
\thanks{Shengjing~Tian is with the School of Economics and Management, China University of Mining and Technology, Xuzhou, China (e-mail:tye.dut@gmail.com), corresponding author.}
\thanks{Yinan Han is with the DUT-BSU Joint Institute, Dalian University of Technology, China (e-mail: Spolico\_hyn@outlook.com).}

%\author{IEEE Publication Technology,~\IEEEmembership{Staff,~IEEE,}
        % <-this % stops a space
%\thanks{This paper was produced by the IEEE Publication Technology Group. They are in Piscataway, NJ.}% <-this % stops a space
%\thanks{Manuscript received October 14, 2022; revised August 13, 2023; revised November 10, %2023; accepted December 17, 2023.}
}

\markboth{Journal of \LaTeX\ Class Files,~Vol.~14, No.~8, August~2021}%
{Shell \MakeLowercase{\textit{et al.}}: A Sample Article Using IEEEtran.cls for IEEE Journals}


\maketitle

\begin{abstract}
3D single object tracking (3DSOT) in LiDAR point clouds is a critical task for outdoor perception, enabling real-time perception of object location, orientation, and motion. 
Despite the impressive performance of current 3DSOT methods, evaluating them on clean datasets inadequately reflects their comprehensive performance, as the adverse weather conditions in real-world surroundings has not been considered. 
One of the main obstacles is the lack of adverse weather benchmarks for the evaluation of 3DSOT.
To this end, this work proposes a challenging benchmark for LiDAR-based 3DSOT in adverse weather, which comprises two synthetic datasets (KITTI-A and nuScenes-A) and one real-world dataset (CADC-SOT) spanning three weather types: rain, fog, and snow. 
Based on this benchmark, five representative 3D trackers from different tracking frameworks conducted robustness evaluation, resulting in significant performance degradations. 
This prompts the question: What are the factors that cause current advanced methods to fail on such adverse weather samples?
Consequently, we explore the impacts of adverse weather and answer the above question from three perspectives: 1) target distance; 2) template shape corruption; and 3) target shape corruption.
Finally, based on domain randomization and contrastive learning, we designed a dual-branch tracking framework for adverse weather, named DRCT, achieving excellent performance in benchmarks.
\end{abstract}


\section{Introduction}

\IEEEPARstart{T}{o} ensure safe autonomous driving, it is essential to have real-time information about the location and movement of objects in the surrounding area, such as other vehicles and pedestrians, no matter what the weather is like. Recently, the LiDAR-based 3D single object tracking (3DSOT) task has been an effective solution for predicting the position and orientation of a target in successive frames using a specified template of an arbitrary target.  
While current 3DSOT methods have achieved impressive performance, they often put attention on sunny scenes but ignore adverse weather conditions. Therefore, it is necessary to systematically evaluate and analyze the robustness of single object tracking methods under different weather conditions.

\begin{figure*}[htb]
  \begin{center}
    \subfloat{
		\includegraphics[scale=0.1]{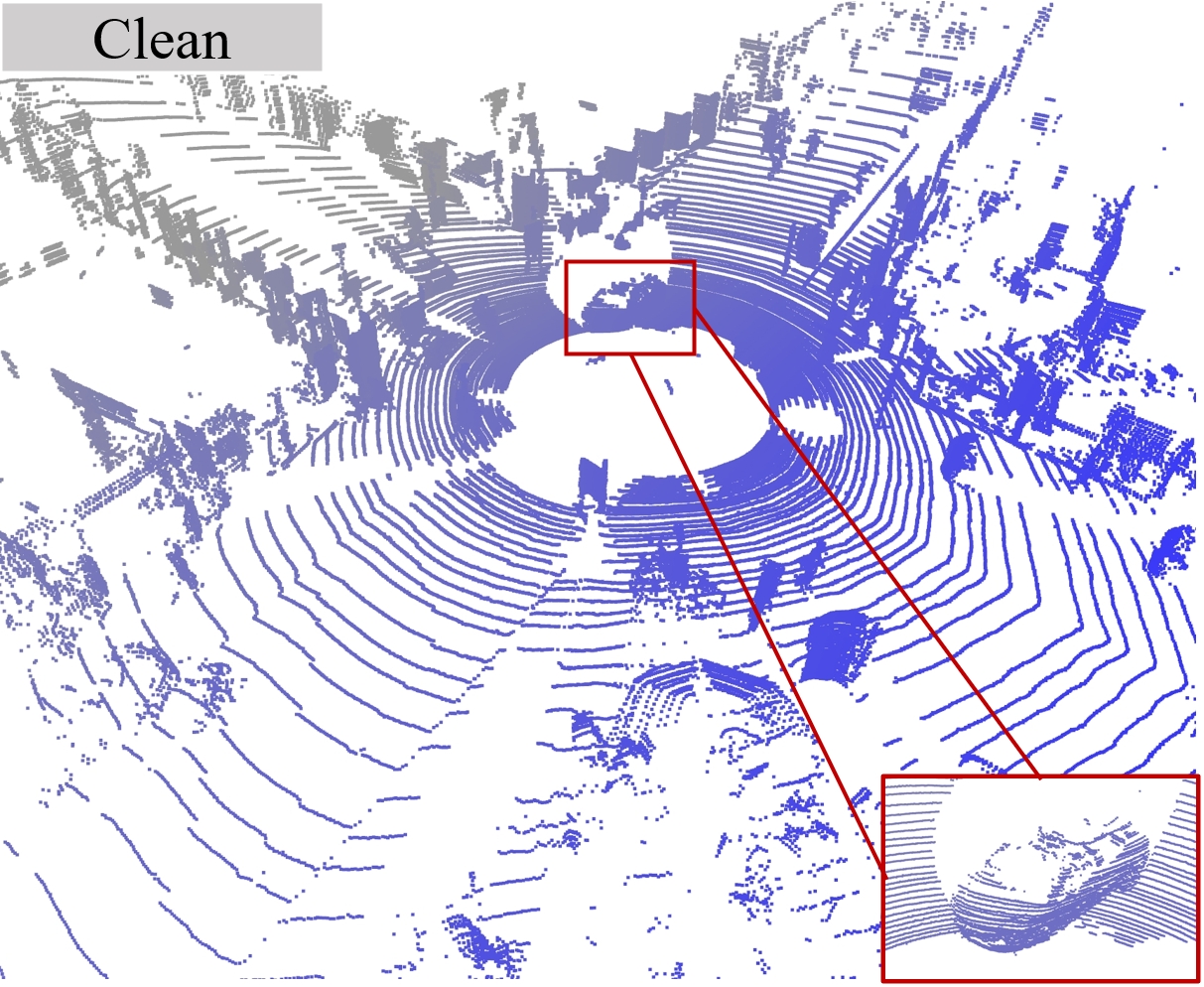}}
    \subfloat{
		\includegraphics[scale=0.1]{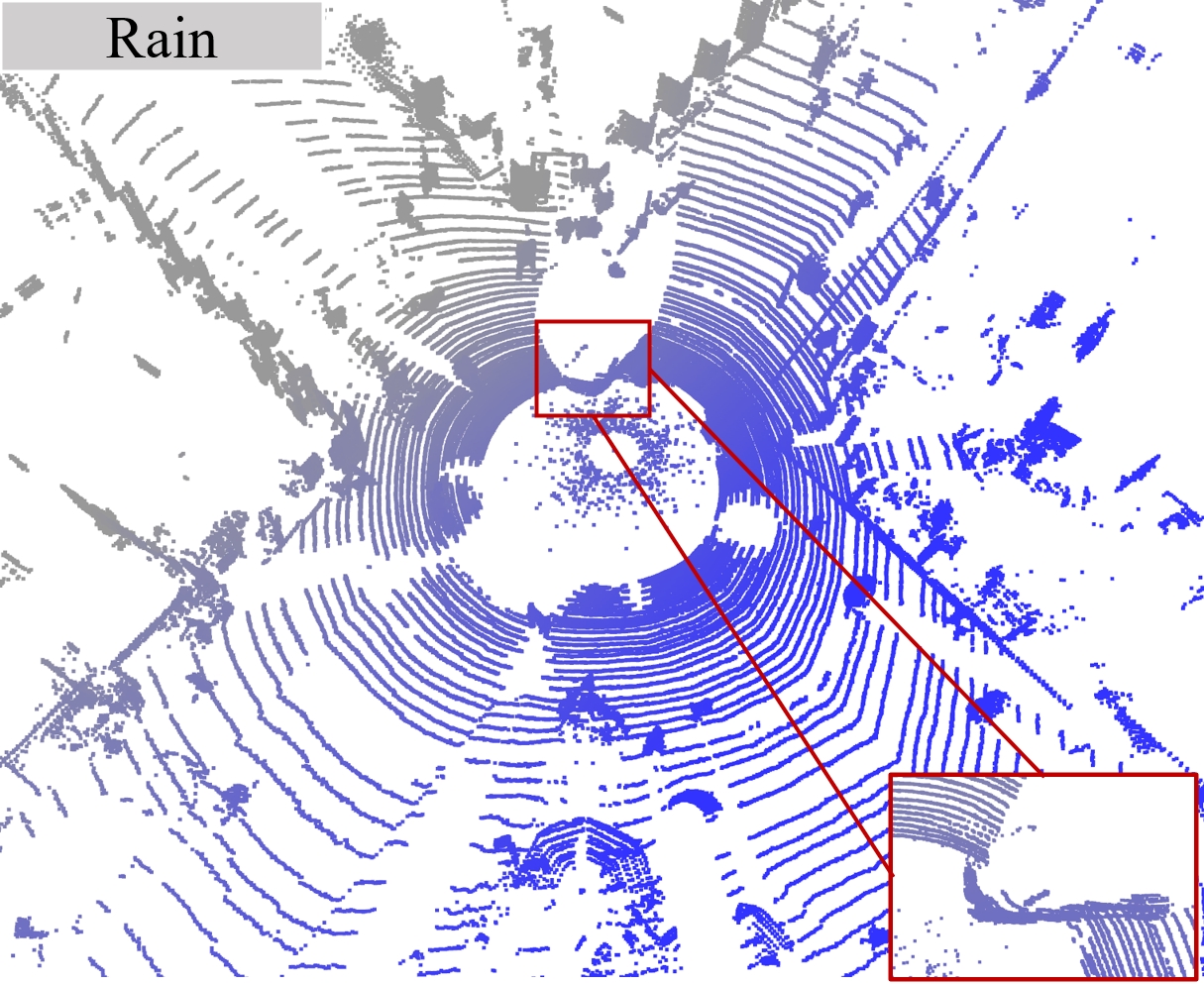}}
    \subfloat{
	\includegraphics[scale=0.1]{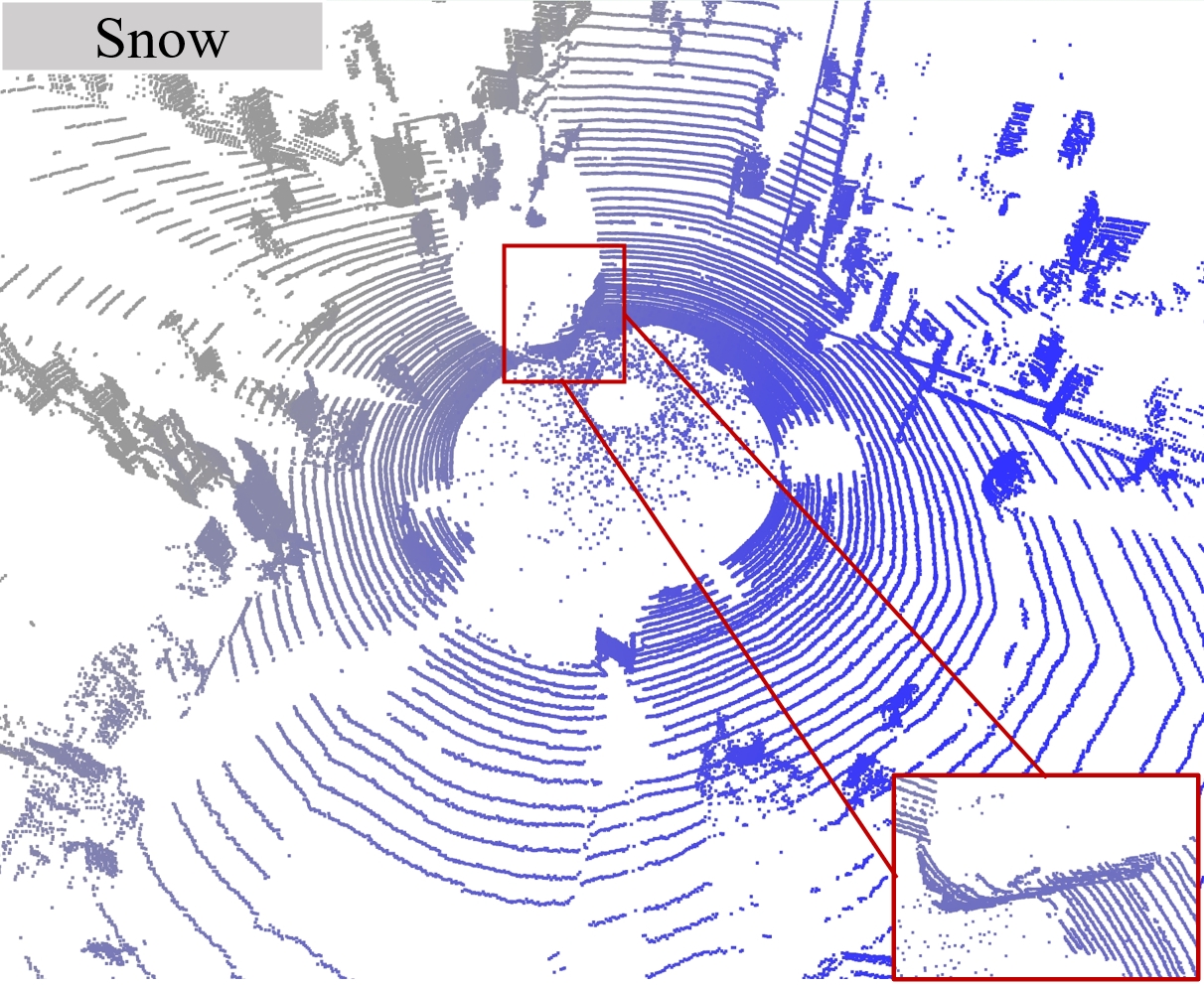}}
    \subfloat{
		\includegraphics[scale=0.1]{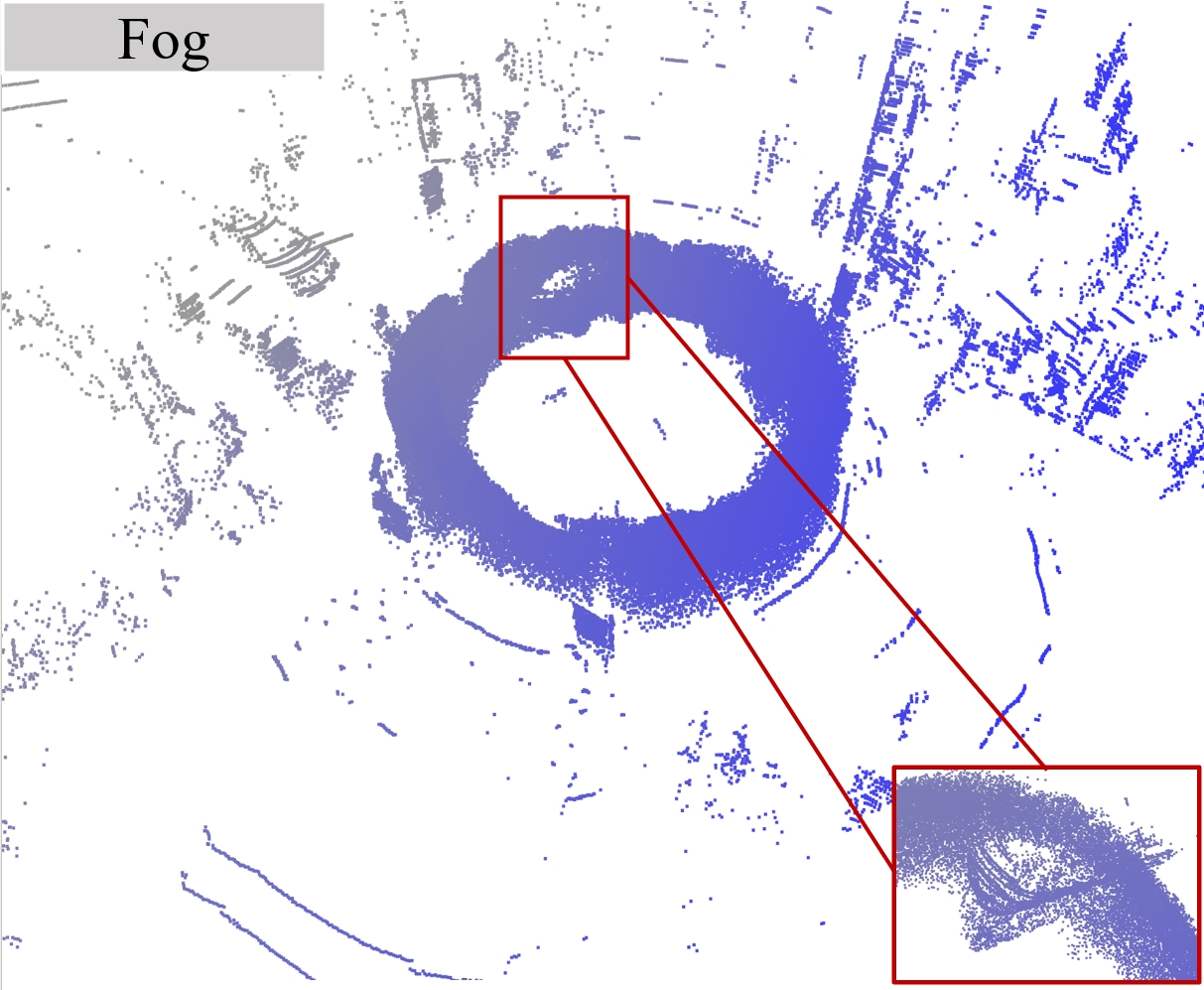}}
  \end{center}
  \caption{Visualization of each weather type in our KITTI-A.} 
  \label{kitti_vis}
\end{figure*}

Currently, state-of-the-art trackers are dedicated to improving accuracy.
Those trackers can be divided into two families: appearance-based and motion-based.
The appearance-based methods extract shape features from both the template and the search area based on a shared backbone and then measure their similarity to model the relationship between them. These trackers extract features via different backbones, such as PointNet++~\cite{PointNet++} (e.g., P2B~\cite{P2B}, V2B~\cite{V2B}, PTTR~\cite{pttr}, OSP2B~\cite{osp2b}, GLT~\cite{glt}), Transformer~\cite{attention} (e.g., STNet~\cite{STNET}, OST~\cite{OST}), and DGCNN~\cite{dgcnn}(e.g., MBPTrack~\cite{mbptrack}, CXTrack~\cite{cxtrack}).
The motion-based methods include MMTracker~\cite{m2track}, SETD~\cite{SETD}, and RDT~\cite{RDT}. MMTracker segments the target in the search area and estimates the positional offset between two frames; RDT aligns the template and search area and computes the rigid transformation; and SETD proposes an accurate state estimator to robustly track target objects in challenging scenarios. Despite the increase in performance over the accuracy metric, there has hardly been any focus on the robustness of the model under adverse conditions.

Robustness research typically involves two aspects. One part focuses on data with minor perturbations and errors ~\cite{imagenet-C,Domain-invariant,adverse_attack,onlyonceattack}, such as those influenced by weather, sensor errors, or adversarial attacks.
The other part concentrates on out-of-distribution (OOD) data, which is usually affected by changes in the acquisition environment~\cite{city2city}, sensor discrepancy~\cite{V2B}, or unknown categories in the open world~\cite{openworld, ood}.
This results in variations in point cloud sparsity, point ranges, object sizes, and unknown objects. Some tentative explorations have been made, but there is still a long way to go in exploring the robustness of 3DSOT tasks. V2B has focused on the robustness for tracking in sparse point cloud scenarios, and DSDM has evaluated the robustness across different datasets and object sizes.
Additionally, recent studies ~\cite{class-agnostic} and~\cite{OST} have shifted their attention to tracking unknown objects towards general 3DSOT.
In spite of the above studies, the 3DSOT task against adverse weather remains untapped.
Therefore, from a safety and stability perspective, there is an urgent need to evaluate the robustness of the trackers under adverse weather conditions.

Evaluating robustness under adverse weather conditions requires corresponding LiDAR point cloud scenes.
Current 3D single object tracking methods are assessed using ``clean'' LiDAR datasets, such as KITTI and nuScenes. However, these datasets lack annotations and scenes captured under adverse weather conditions, making it impossible to evaluate the trackers' robustness on such datasets.
We also note that some LiDAR datasets for adverse weather have recently been published, such as Seeing Through Fog (STF)~\cite{STF}, Canadian Adverse Driving Conditions (CADC)~\cite{cadc}, and Ithaca365~\cite{ithaca365}. Nevertheless, since some lack sequential object information or require additional processing to construct the tracking sequences, they cannot be directly employed for the 3DSOT task.

In this work, we benchmark the robustness of 3D single object trackers in adverse weather conditions.
It comprises two synthetic datasets (KITTI-A and nuScenes-A) and one real-world dataset (CADC-SOT) spanning three weather types: rain, fog, and snow. To provide a more realistic evaluation, we generate synthetic datasets through weather simulations and also manipulate a real-world dataset collected during snowfall. Specifically, for the synthetic datasets, we leverage the adverse weather simulation algorithms to incorporate the effect of rain, snow, and fog into the clean datasets KITTI~\cite{kitti} and nuScenes~\cite{nuScenes}. These datasets are chosen based on different point densities resulting from their collection methods: KITTI with 64-beam LiDAR and nuScenes with 32-beam LiDAR. Each type of adverse weather condition includes five levels of intensity within both KITTI-A and nuScenes-A, allowing us to examine how various sensor resolutions handle these challenges. As for the real-world dataset, we perform target filtration and snow weather classification based on the CADC dataset \cite{cadc} to construct effective tracking segments. Our CADC-SOT dataset is composed of four categories of deteriorated data, encompassing three gradations of snowfall as well as the accumulation of snow covering road surfaces. This enables us to compare the performance of different trackers under actual adverse weather conditions.

Based on the benchmark, we explore the impacts of adverse weather on object tracking from three perspectives: 1) target distance; 2) template shape corruption; and 3) target shape corruption. We find that current tracking methods suffer a significant degradation of tracking performance in adverse weather, which is explained by the fact that such weather alters the distribution of the captured point clouds, making the a prior of such data absent. How to improve the tracking performance in normal weather while guaranteeing the performance in such adverse weather is then a key issue. 
To this end, we propose a dual-branch training method based on domain-randomized contrast learning to enhance the robustness of the model. The core techniques consist of assist branch-based domain randomized enhancement strategy and local geometric contrast module (LGCM). Specifically, the primary branch accepts clean data as input, while the auxiliary branch feeds with randomized point cloud data enhanced by perturbation, point dropping, and point adding. After feature extraction, the LGCM empowers the perception capabilities from the random domain to the primary branch through central contrast constraints. 

In this work, we contribute in four key aspects:
\begin{itemize}
    \item We present three datasets representing adverse weather conditions to assess the robustness and stability of 3D single object tracking methods, encompassing rainy, foggy, and snowy conditions, along with both synthetic and real-world datasets.
    
    \item We conduct a thorough examination of existing methods on our proposed benchmark, evaluating the robustness of various architectures.
    
    \item We conducted a comprehensive analysis of the impact of adverse weather conditions on 3D single object tracking, taking into account factors such as target distance, template quality, and the quality of the tracking target. This has laid the groundwork for subsequent work and provided valuable insights for future research directions.

    \item We designed a robust tracking method for adverse weather, named DRCT, which not only improves the performance of the tracking algorithm under clean weather but also achieves robust tracking in adverse conditions, resulting in performance enhancement in our benchmarks.
    
\end{itemize}

\section{Related Works}

\begin{figure*}[htb]
  \begin{center}
    \subfloat{
		\includegraphics[scale=0.1]{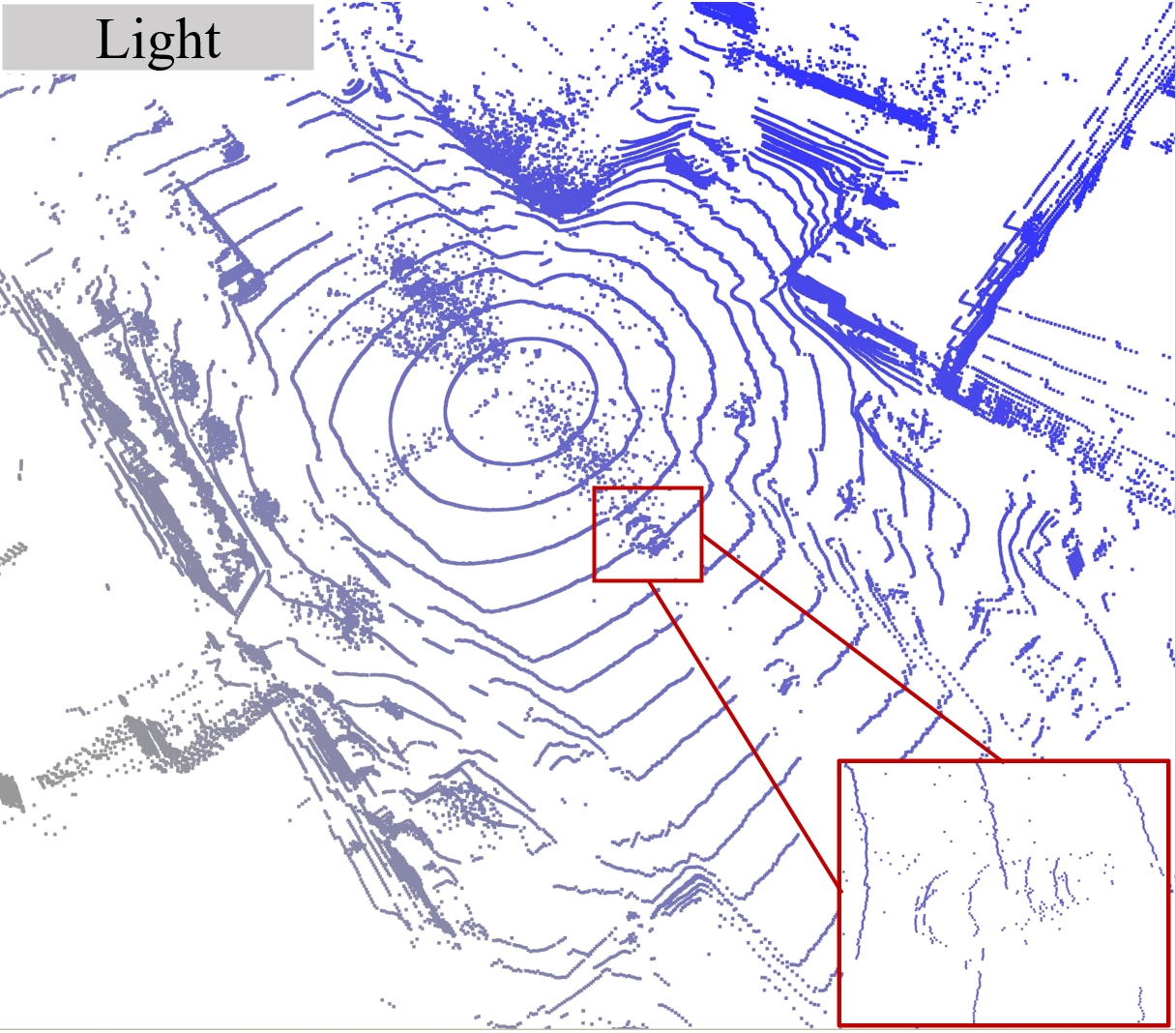}}
    \subfloat{
		\includegraphics[scale=0.1]{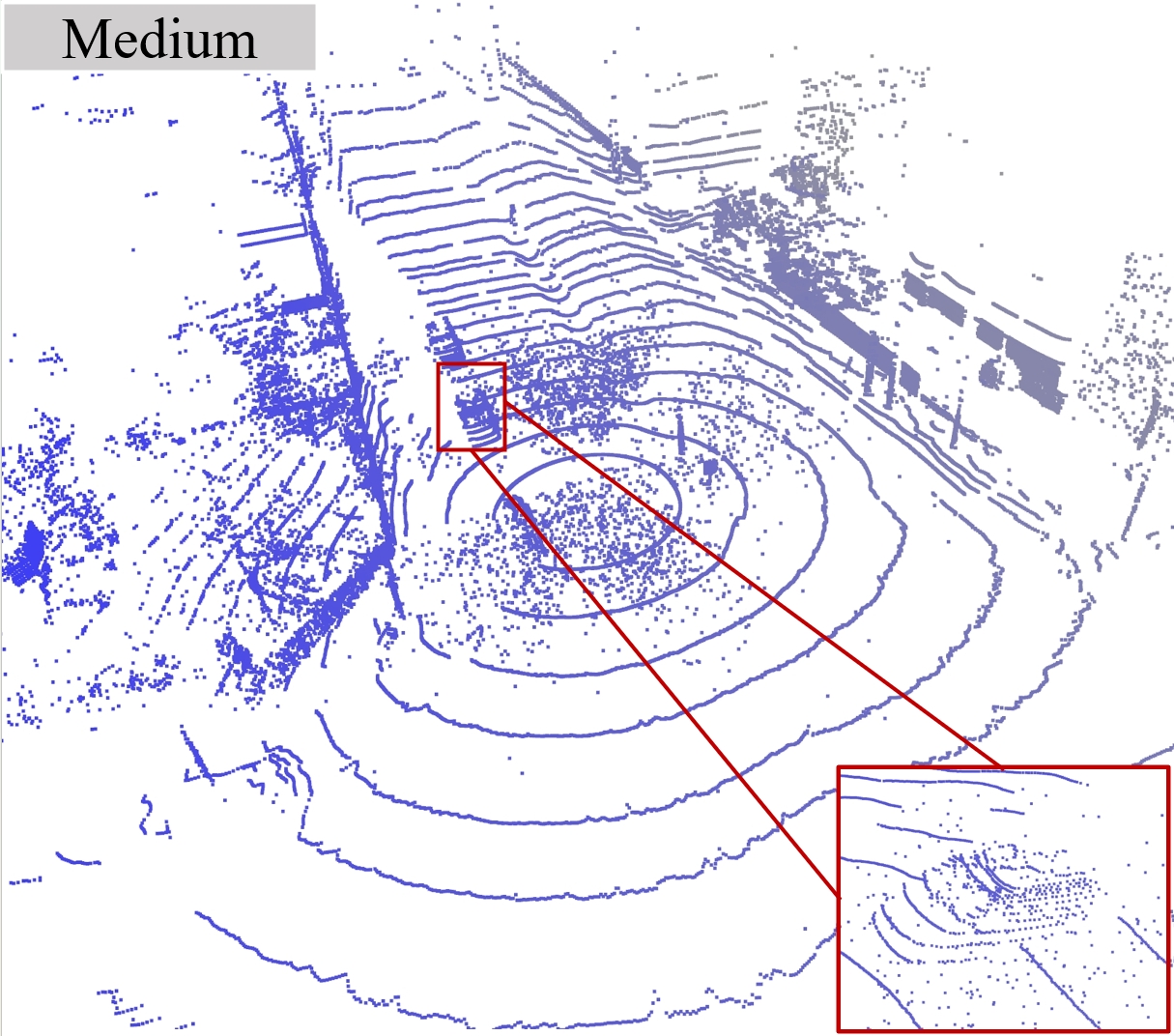}}
    \subfloat{
	\includegraphics[scale=0.1]{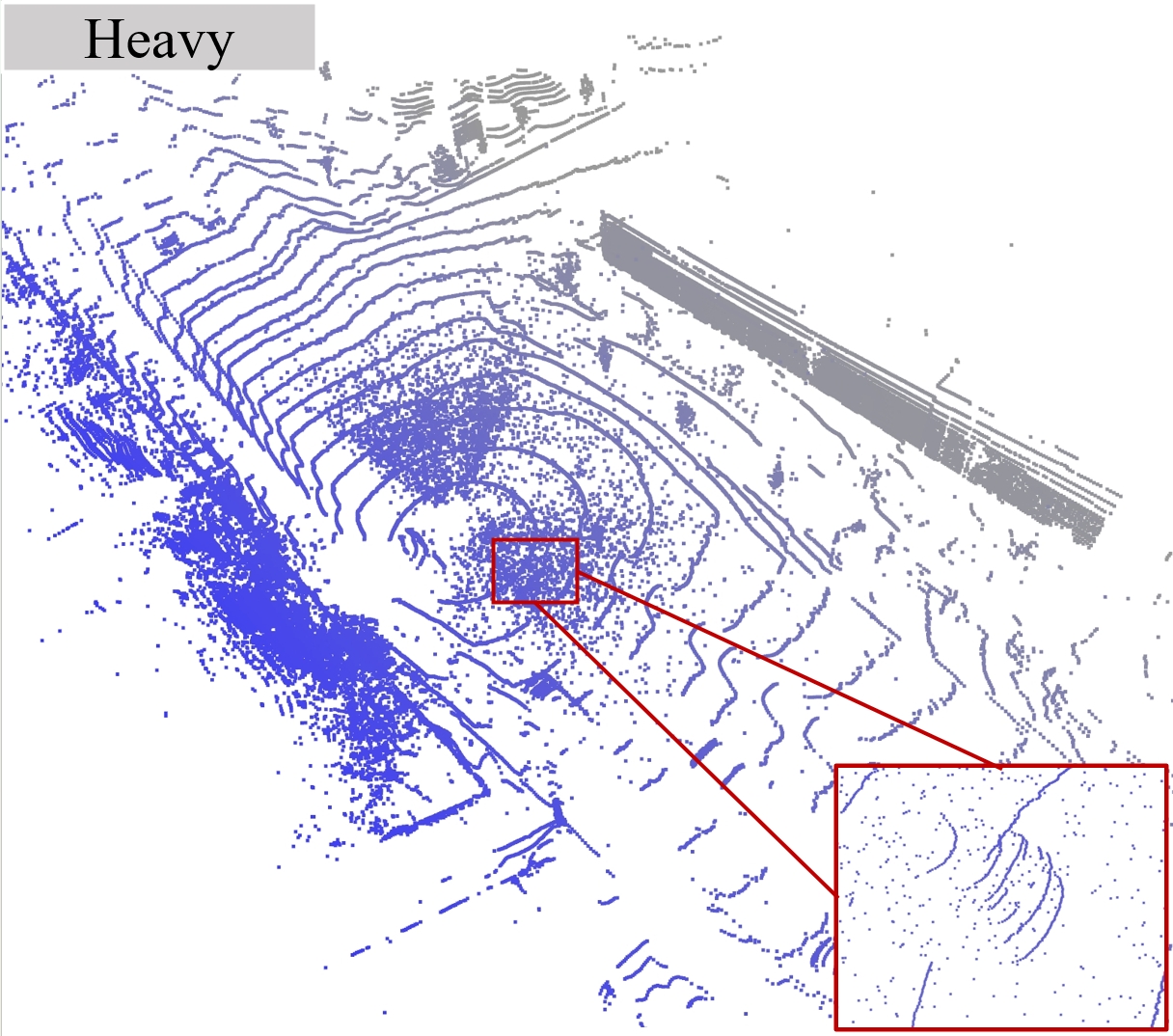}}
    \subfloat{
		\includegraphics[scale=0.1]{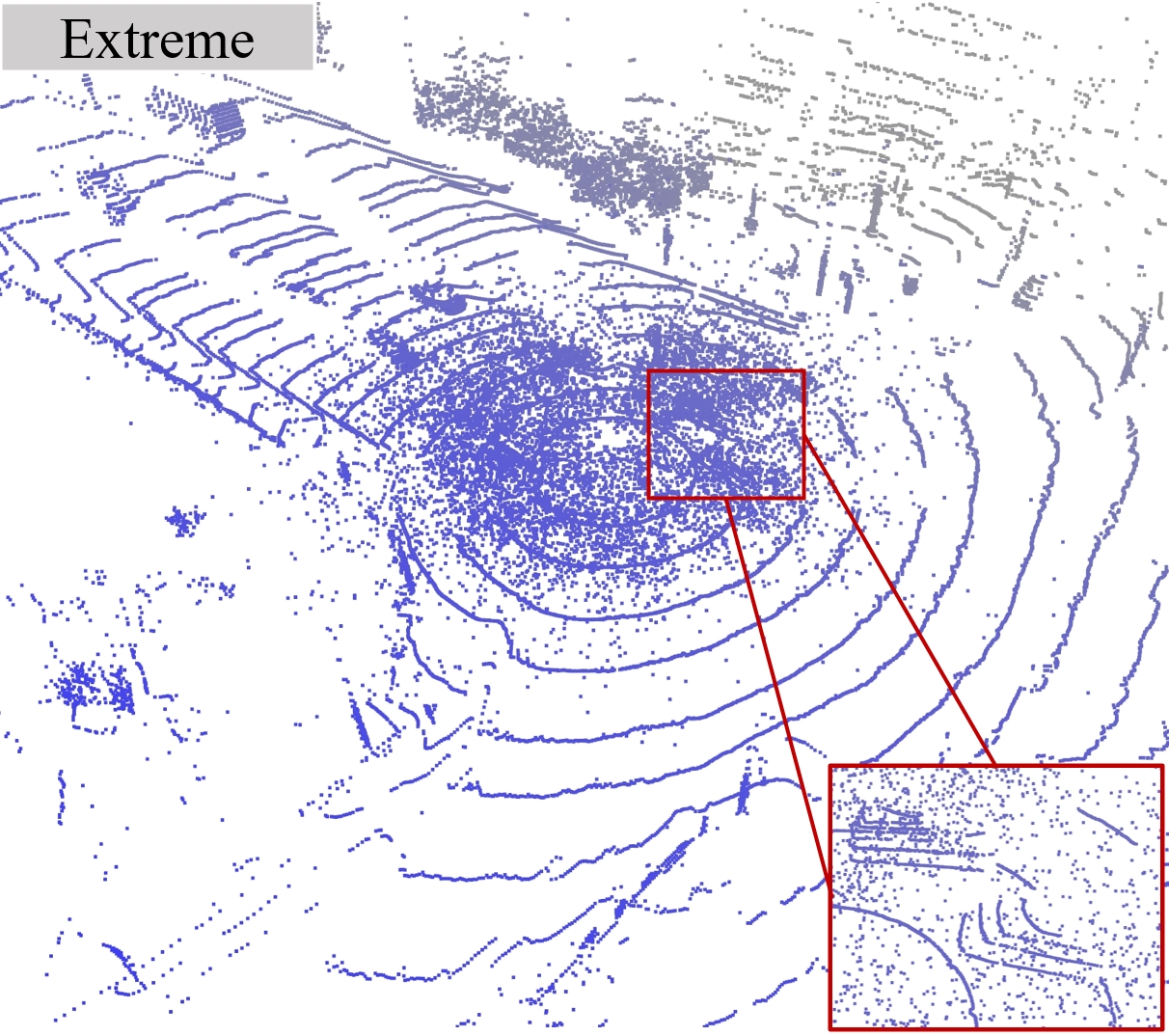}}
  \end{center}
  \caption{Visualization of each level snowy in our CADC-SOT.} 
  \label{cadc_vis}
\end{figure*}

\subsection{3D Single Object Tracking}

Recently, 3D single object trackers have been achieving considerable performance on clean LiDAR datasets. These trackers can be classified into appearance-based trackers~\cite{BAT,P2B,LTTR,mbptrack,cxtrack,pttr,V2B,STNET,PTT,OST}  and motion-based trackers~\cite{m2track,DSDM,RDT}. 
SC3D~\cite{sc3d} was the first to propose a method for single object tracking via appearance matching.
P2B~\cite{P2B} introduced a Siamese network for point clouds, summarizing 3D single object tracking as feature extraction, relationship modeling, and target prediction.
To address problems with objects of different sizes but similar scan patterns, BAT~\cite{BAT} introduced geometry features with Box Prior.
PTTR~\cite{pttr} used a Transformer to establish relationships between the template and search region, employing a two-stage head for target state prediction.
V2B~\cite{V2B} enhanced tracking in sparse scenes with a local-global relationship model and target state estimation on the Bird's Eye View (BEV).
Based on V2B, STNet~\cite{STNET} utilized Transformer modules for feature extraction and relationship modeling, further enhancing performance.
CXTrack~\cite{cxtrack} considered contextual relationships between consecutive frames, while MBPTrack~\cite{mbptrack} adopted a memory mechanism to leverage past information, formulating localization in a coarse-to-fine scheme using Box Priors.

The motion-based tracker MMTrack~\cite{m2track} segmented the target point cloud to predict motion between frames, enabling target state refinement.
However, they overlooked robustness and stability. In this work, we aim to evaluate performance in adverse weather conditions and identify current tracker shortcomings.

\subsection{Adverse Weather Datasets}

Adverse weather conditions constitute prevalent and indispensable scenarios in the context of autonomous driving. Several datasets tailored for adverse weather environments have been proposed, primarily oriented towards advanced tasks such as 3D segmentation~\cite{segbenchmarking},\cite{semaniticstf},\cite{robo3d} and detection~\cite{detbenchmarking}.
These works can be divided into two aspects, the first of which is the collection of real data on adverse weather conditions.
For example, the Seeing Through Fog (STF)~\cite{STF}, Canadian Adverse Driving Conditions (CADC)~\cite{cadc}, and Ithaca365 datasets~\cite{ithaca365} are collected under the adverse weather. However, they are designed for 3D detection; they are not suitable for other tasks. Although SemanticSTF~\cite{semaniticstf} has labeled the point-wise annotations on STF, it still lacks the continuous frame annotation used for tracking tasks. CADC is collected on snowy days and has temporal labels. However, further screening is required due to sparse point clouds and target occlusion.

The second of which is to synthesize real-world corruption on clean datasets. For example, in image classification, ImageNet-C~\cite{imagenet-C} is first introduced with 15 corruption types, ranging from noise, blur, and adverse weather to digital corruptions. In 3D point cloud detection, there are KITTI-C, nuScenes-C, and WayMo-C~\cite{detbenchmarking} with 27 types of common corruptions for both LiDAR and camera inputs. In 3D semantic segmentation, SemanticKITTI-C~\cite{segbenchmarking} is to comprehensively analyze the robustness of LiDAR semantic segmentation models under 16 types of corruptions. In 3D perception, Robo3D~\cite{robo3d} presented a comprehensive benchmark heading toward probing the robustness of 3D detectors and segments under OOD scenarios against natural corruptions. Authentic data reflects the real-world environment, while simulated data can encapsulate increasingly comprehensive scenarios. In this study, we establish benchmarks suitable for 3D object tracking, encompassing both simulation and real-world data.

\subsection{Robustness Study in LiDAR Point Clouds}

In real-world applications, LiDAR point clouds are inevitably subject to robustness issues. This can be summarized as two research aspects. Firstly, there is a focus on robustness to unknown data from out-of-distribution domains. This includes scenarios like transitioning from simulation to reality~\cite{sim2real}, day to night~\cite{day2night}, one city to another~\cite{city2city}, and known to unknown category~\cite{class-agnostic,OST}. These studies provide valuable references and inspiration for 3D SOT tasks.
The second aspect is to ensure robustness to irregular data within the same distribution domain.
In this regard, some methods assessed the robustness of the model by simulating or collecting corrupted data, such as data affected by adverse weather \cite{STF,lisa,ithaca365,fog,semaniticstf,segbenchmarking}, sensor errors \cite{robo3d}, and fast movements \cite{detbenchmarking}. Some methods utilized adversarial attacks to examine the robustness by purposely designing small imperceptible perturbations \cite{attack,attack_1}. 
While these methods in point cloud tasks such as detection, segmentation, and completion can provide relevance, there is still no relevant work in 3DSOT tasks.
In this study, we aim to explore method robustness to irregular data using a designed dataset affected by adverse weather.

3D perception under adverse conditions has gained significant attention in the industry, and with the introduction of several perception task benchmarks, methods to enhance perception robustness can be categorized into two approaches: data augmentation and feature enhancement. 
For the former, some methods\cite{denoising, PointFilterNet} eliminate the impact of adverse weather by denoising point clouds, while others enable the model to adapt to unfavorable conditions through weather simulation\cite{fogsim, snowsim}.
Furthermore, there are studies \cite{rethinkdataaug} that improve model robustness via data augmentation methods to tackle the challenges presented by adverse weather.
For the latter, the representational capacity of the network is a key factor determining model robustness. Some studies \cite{robo3d,unimix,semaniticstf} enhance the expressive power of model features by transferring knowledge through knowledge distillation or contrastive learning. However, these methods focus too much on robustness under adverse conditions and fail to ensure performance under normal conditions. 
We employed domain randomization techniques to augment point cloud data as input. The assist branch is trained on disturbed random domains, while the main branch uses random domain knowledge to assist training under normal conditions. This strategy ensures the model's performance under normal conditions while enhancing its robustness in adverse conditions.

\section{Benchmark}

To assess the robustness of existing 3D target object tracking algorithms, we devised a benchmark under the setting of adverse weather conditions. These benchmarks encompass real-world adverse weather scenarios as well as extensive simulation scenarios to provide a comprehensive evaluation.
%问题定义
\subsection{Problem} 
In 3D single object tracking, the fundamental goal of a tracker is to precisely determine the spatial location of a specified target within each frame of a given point cloud sequence. Typically, the target is initially defined in the first frame, serving as a reference to guide the subsequent tracking process. Mathematically, when provided with a template $P_t$ and a search region $P_s$, comprised of $N_t$ and $N_s$ coordinates $(x,y,z)$ respectively, a tracker is supposed to accurately forecast the state of the designated object within the search region $P_s$. For the representation of the target's state, we employ a 3D bounding box, parameterized by its center $(x,y,z)$, sizes $(l,w,h)$, and yaw angle $\theta$. Given that outdoor scenes captured by LiDAR typically preserve the physical size of rigid targets, the tracking problem can be formalized as follows:

\begin{equation}\label{tracker}
\begin{aligned}
   \{x,y,z,\theta \} = \Phi(P_t, P_s),
\end{aligned}
\end{equation}
where $\Phi$ represents the tracking model. 

\subsection{Synthetic Data Generation}

\begin{table}[htb]
  \caption{Statistic on KITTI-A and nuScene-A. The number of target sequences and points cloud frames are reported.}
  \label{tab:syn_data}
  \centering
   \scalebox{0.80}{
  \setlength{\tabcolsep}{1.0mm}{
  \begin{tabular}{l|cc|cc}
    \toprule
       & \multicolumn{2}{c}{Car}&  \multicolumn{2}{c}{Pedestrian} \\
       & Sequences & Frames &  Sequences & Frames   \\ 
    \midrule
    KITTI-A &1,800 & 96,360 & 930& 91,320\\
    nuScenes-A & 54,900 &977,385 &27,870 &498,405 \\
    \bottomrule
  \end{tabular}}
  }
\end{table}

The tracking problem described above claims that the tracker needs to produce enclosed bounding boxes that are temporally coherent when localizing targets. While some existing 3D detection datasets \cite{detbenchmarking} simulate point distributions in adverse weather conditions using single-frame scanned data, they are not suitable for tracking time-varying objects in point cloud videos.

In this work, we generated the synthetic data with temporal information. Furthermore, we instituted two benchmarks, utilizing the tracking split of KITTI and nuScenes.
To enhance the realism of these datasets, we incorporated weather simulations, encompassing a range of conditions, including fog, rain, and snow, into their validation sets. Each of these weather conditions was further subdivided into five distinct levels, giving rise to our KITTI-A and nuScenes-A datasets. In Table \ref{tab:syn_data}, we report the number of valid sequences and frames of cars and pedestrians. These samples are 15 times larger than the original dataset because we did 15 weather simulations (3 types of weather, 5 levels) for each class in each dataset.
Taking the cars of KITTI as examples, the original data is 6,424 frames, and the number of frames after weather simulations is $3\times5\times6,424 = 96,360$. We visualized it in Figure. \ref{kitti_vis}.

Specifically, to simulate foggy scenarios, we employed the methodology outlined in \cite{fog}. The parameter responsible for controlling the meteorological optical range was meticulously set at five discrete values. This enabled us to represent varying levels of fog with varying degrees of density, ranging from a light mist to a dense fog. For rainy and snowy conditions, we employed the LISA~\cite{lisa} simulation method for KITTI-A and nuScenes-A, setting parameters such as rainfall rate and snowfall rate to simulate different levels. In this way, the synthetic data significantly enriches and completes the benchmark, effectively compensating for variations across different devices and collection locations. This strategic data augmentation substantially facilitates a more thorough and comprehensive evaluation of the trackers' robustness.

\begin{table}[htb]
  \caption{Statistic on CADC-SOT. The table shows the number of scenarios and frames in The table shows the number of scenarios and frames in different snow levels and whether snow is covered on the road}
  \label{tab:CADC-SOT-info}
  \centering
   \scalebox{0.80}{
  \setlength{\tabcolsep}{1.0mm}{
  \begin{tabular}{l|cc|cc}
    \toprule
     \multirow{2}{*}{CADC-SOT} & \multicolumn{2}{c}{Car}&  \multicolumn{2}{c}{Pedestrian} \\
       & sequences & frames&  sequences & frames   \\
       
    \midrule  
    Light &1,173 & 33,120&365 &6,855 \\
    Medium & 1,960&41,166 &477 &10,068 \\
    Heavy & 1,423&26,496 & 179&4,253 \\
    Extreme & 1,048&19,282 &150 &2,350 \\
    \midrule 
    Covered & 4,485&87,854 & 884&18,054 \\
    Non covered & 1,719&32,210 &287 &5,472 \\
    \midrule 
    All & 6,204&120,064 &1,171 &23,526 \\
    \bottomrule
  \end{tabular}}
  }
\end{table}

\subsection{Realistic Data Generation}

Collecting data in real weather conditions can provide a more accurate representation.
The CADC dataset \cite{cadc} is an extensive collection of LiDAR data, meticulously captured during genuine snowy days. It encompasses four distinct levels of snowfall intensity (light, heavy, medium, and extreme) and comprises 10 distinct categories of objects. This dataset is unique when compared to \cite{ithaca365,STF} in that it offers not only LiDAR scans in snow weather but also includes temporal information for tracking.

In this work, the proposed CADC-SOT is generated by filtering invalid targets and grading snow weather based on the CADC dataset. We first removed the frames with less than 10 points of target in the original CADC sequences, then we picked out the remaining sub-sequences of length greater than 4. In addition, we also divided the selected sequences into two groups based on whether roads are covered by snow to test the ground impact for tracking. 
As shown in Table~\ref{tab:CADC-SOT-info}, our CADC-SOT consists of 7,375 tracking sequences, including 6,204 car sequences and 1,171 pedestrian sequences. The car sequences have 120,064 frames, and the pedestrian sequences have 23,526 frames. We visualized it in Figure. \ref{cadc_vis}.

\subsection{Metric}
Robustness reflects the extent of degradation in the tracker's performance under adverse weather conditions. To quantify this change more clearly, we employ the degradation rate ($DR$) as a relative robustness metric for measuring how much accuracy a model can retain when evaluated on the corruption sets. The $DR$ can be written as

\begin{equation}
   \begin{aligned}
    DR^{S}_{i,j} &= \frac{S_{i,j}} {S_{c}}, \\
    \end{aligned}
\end{equation}

where $S\in\{success, precision\}$ is the metric; $S_{c}$ represents the performance on a clean dataset; and $i,j$ is the type and level of adverse weather, respectively. Additionally, based on $success$ and $precision$ metrics, we also quantify the stability of the method using range ($R$) and standard deviation ($S.d$) across various levels. Here, the range ($R$) is calculated by subtracting the minimum value from the maximum value.

\begin{table*}[htb]
 \centering
  \caption{The results of different trackers on KITTI-A benchmark. The table documents the tracking success/precision of each method for various levels of adverse weather, along with the average degradation rate($DR$), range($R$), and standard deviation($S.d$). The best results are highlighted in bold.}
  \label{tab:KITTI-results}
  \begin{tabular}{c|c|c|ccccc|ccc}
    \toprule
       \multirow{2}{*}{~} & \multirow{2}{*}{Method} & \multirow{2}{*}{Clean} & \multicolumn{5}{c|}{KITTI-A}& \multirow{2}{*}{$DR$} & \multirow{2}{*}{$R$}  & \multirow{2}{*}{$S.d$} \\
      & &  & Lv-1 & Lv-2 & Lv-3 & Lv-4 & Lv-5 & ~ & ~ & ~ \\ 
    \midrule  
       \multicolumn{11}{c}{$\mathbf{Rain}$}\\
    \midrule
        \multirow{5}{*}{Car} 
         & BAT &  62.83/75.34 & 36.41/43.93 & 36.82/43.88 & 36.73/44.10 & 36.08/43.27 & 37.41/44.74 & 0.42/0.42 & 1.33/\textbf{1.47}  & 0.50/\textbf{0.53} \\
         & MMTrack & 67.43/81.04 & 43.33/54.53 & 42.29/52.94 & 42.74/53.59 & 42.76/53.51 & 42.27/53.06 & 0.37/0.34 & \textbf{1.06}/1.59  &\textbf{ 0.43}/0.63 \\
         & STNet  & 67.29/78.76 & 38.50/44.98 & 38.87/45.60 & 41.49/48.90 & 41.00/48.20 & 39.11/45.87 & 0.41/0.41 & 2.99/3.92  & 1.35/1.73 \\
         & CXTrack & 69.16/81.66 & 43.15/51.01 & 45.26/53.57 & 46.78/55.56 & 45.08/53.24 & 45.11/53.35 & 0.35/0.35 & 3.63/4.55  & 1.29/1.61 \\
         & MBPTrack & 73.70/85.22 & \textbf{49.28}/\textbf{58.94} & \textbf{48.32}/\textbf{57.58} & \textbf{46.21}/\textbf{55.11} & \textbf{51.11/61.36} & \textbf{48.64/57.83} & \textbf{0.34/0.32} & 4.90/6.25  & 1.77/2.27 \\

    \cmidrule(lr){2-11}
        \multirow{5}{*}{Ped} 
        &  BAT & 39.95/65.40 & 42.08/68.37 & 45.30/74.34 & 40.94/69.81 & 40.22/62.79 & 42.11/73.87 & \textbf{-0.05/-0.07} & 5.08/11.55  & 1.94/4.70 \\
         & MMTrack & 60.60/89.37 & 58.48/87.77 & 59.24/88.45 & 58.39/87.59 & 58.50/87.72 & 58.86/87.96 & 0.03/0.02 & \textbf{0.85}/\textbf{0.86}  & \textbf{0.35/0.34} \\
         & STNet & 47.17/72.99 & 43.45/69.15 & 46.03/72.44 & 45.59/71.83 & 44.71/69.80 & 45.35/71.36 & 0.05/0.03 & 2.58/3.29  & 1.00/1.39\\
         & CXTrack & 65.21/89.39 & 54.65/80.36 & 60.05/82.76 & 58.49/82.65 & 58.58/83.15 & 57.13/80.97 & 0.11/0.08 & 5.40/2.79  & 2.03/1.23 \\
         & MBPTrack & 66.00/90.92 & \textbf{65.26/89.49 }& \textbf{65.42/90.89} & \textbf{65.91/90.88} & \textbf{64.31/88.64} & \textbf{67.40/92.46} & 0.005/0.004 & 3.09/3.82  & 1.13/1.47 \\
        \midrule 
       \multicolumn{11}{c}{$\mathbf{Fog}$}\\
        \midrule 
        \multirow{5}{*}{Car} 
        &  BAT & 62.83/75.34 & 52.93/64.59 & 47.69/59.15 & 49.25/58.72 & 48.69/59.55 & 48.72/57.01 & 0.21/0.21 & 5.24/7.58  & 2.02/2.85\\
         & MMTrack & 67.43/81.04 & 55.81/67.80 & 55.71/\textbf{67.62} & 55.49/\textbf{67.47} & 55.32\textbf{/67.43} & 55.27/\textbf{67.39} & \textbf{0.18/0.17 }& \textbf{0.54/0.41}  & \textbf{0.24/0.17}  \\
         & STNet & 67.29/78.76 & 53.70/63.93 & 52.10/61.94 & 51.78/61.65 & 58.87/70.42 & 51.64/61.53 & 0.20/0.19 & 7.23/8.89  & 3.05/3.78  \\
         & CXTrack & 69.16/81.66 & 57.50/69.25 & 51.20/62.21 & 51.01/62.13 & 50.84/61.90 & 50.95/62.02 & 0.24/0.22 & 6.66/7.35  & 2.91/3.22 \\
         & MBPTrack & 73.70/85.22 & \textbf{59.28/70.13} & \textbf{56.70}/66.98 & \textbf{56.39}/66.64 & \textbf{56.41}/66.71 & \textbf{56.32}/66.56 & 0.23/0.21 & 2.96/3.57  & 1.27/1.53\\

    \cmidrule(lr){2-11}
        \multirow{5}{*}{Ped} 
        &  BAT & 39.95/65.40 & 25.11/43.34 & 16.95/33.27 & 14.56/25.40 & 15.01/28.97 & 15.07/30.54 & \textbf{0.57/0.51} & \textbf{10.55/17.94}   & \textbf{4.44/6.80} \\
         & MMTrack & 60.60/89.37 & \textbf{35.09/57.89} & 23.36/39.40 & \textbf{16.62/31.25} & \textbf{17.62/32.74} & \textbf{17.27/32.22}\textbf{ }& 0.64/0.57  & 18.47/26.64   & 7.81/11.2  \\
         & STNet & 47.17/72.99 & 28.84/49.91 & 19.74/35.32 & 15.56/27.93 & 15.62/29.36 & 16.67/31.02 & 0.59/0.53  & 13.28/21.98   & 5.60/8.94 \\
         & CXTrack & 65.21/89.39 & 32.53/55.10 & 21.02/36.68 & 14.81/29.11 & 14.73/28.68 & 14.94/28.93 & 0.70/0.60  & 17.80/26.42   & 7.71/11.36  \\
         & MBPTrack & 66.00/90.92  & 33.63/54.12 & \textbf{25.30/42.66} & 13.95/27.68 & 14.46/27.72 & 14.09/27.58 & 0.69/0.60  & 19.68/26.54   & 8.88/12.06  \\
        \midrule
       \multicolumn{11}{c}{$\mathbf{Snow}$}\\
        \midrule 
        \multirow{5}{*}{Car} &  
        BAT & 62.83/75.34 & 37.34/44.40  & 36.35/43.54 & 34.88/41.55 & 36.07/43.71 & 35.37/42.31 & 0.43/0.43 & 2.46/2.85  & 0.94/1.15 \\
         &MMTrack & 67.43/81.04 & 43.20/53.96 & 42.74/53.51 & 42.61/53.40 & 42.14/52.83 & 42.06/52.43 & 0.37/\textbf{0.34 }&\textbf{1.14/1.53}  & \textbf{0.47/0.60} \\
         &STNet & 67.29/78.76 & 41.56/48.97 & 41.39/48.74 & 38.49/45.04 & 40.56/47.42 & 39.71/46.41 & 0.40/0.41 & 3.07/3.93  & 1.27/1.64 \\
         &CXTrack & 69.16/81.66 & 44.72/52.68 & 45.55/53.84 & 43.90/51.90 & 43.34/51.07 & 39.65/46.03 & 0.37/0.37 & 5.90/7.81 & 2.27/3.01 \\
         &MBPTrack & 73.70/85.22 & \textbf{49.19/58.71} & \textbf{48.11/57.06} & \textbf{48.27/57.45} & \textbf{45.74/54.27} & \textbf{46.09/54.94 }& \textbf{0.36/0.34} & 3.45/4.44  & 1.49/1.84 \\
  
    \cmidrule(lr){2-11}
        \multirow{5}{*}{Ped} 
         & BAT & 39.95/65.40 & 40.09/67.22 & 44.11/72.10 & 43.67/70.97 & 42.37/70.78 & 42.32/70.00 &\textbf{ -0.06/-0.07} & 4.02/4.88  & 1.57/1.83 \\
         & MMTrack & 60.60/89.37 & 57.81/86.88 & 58.80/87.97 & 58.63/87.91 & 56.35/85.06 & 57.79/86.80 & 0.05/0.03 & 2.45/\textbf{2.91}  & 0.97/1.19 \\
         & STNet & 47.17/72.99 & 47.76/75.45 & 44.61/68.91 & 46.31/72.53 & 43.58/68.15 & 47.51/74.95 & 0.03/0.01 & 4.18/7.30  & 1.82/3.36\\
         & CXTrack & 65.21/89.39 & 62.32/85.41 & 58.47/83.31 & 64.54/89.13 & 57.43/81.36 & 58.44/82.75 & 0.08/0.06 & 7.11/7.77  & 3.04/3.02 \\
         & MBPTrack & 66.00/90.92 & \textbf{66.78/91.53} & \textbf{65.40/90.22} & \textbf{65.34/90.20} & \textbf{64.87/89.29} &\textbf{64.72/88.60} & 0.01/0.01 &\textbf{2.06}/2.93  &\textbf{0.81/1.11} \\
    \bottomrule
  \end{tabular}
\end{table*}

\section{Evaluation}

To evaluate the performance of advanced methods, we conducted robustness evaluation from two aspects: real-world scenes and synthetic scenes. Five representative tracking methods are selected to assess their degradation rate of tracking metrics, including BAT~\cite{BAT}, MMTrack~\cite{m2track}, MBPTrack~\cite{mbptrack}, STNet~\cite{STNET}, and CXTrack~\cite{cxtrack}. In addition, we measure the stability of the method in terms of range and standard deviation, i.e., fluctuations in performance under different conditions.

\begin{figure*}[htb]
    \centering
    \rotatebox{90}{~~~~~~~~~~~~Target Distance}
    \begin{minipage}[t]{0.32\linewidth}
    \centering
    \includegraphics[width=1\linewidth]{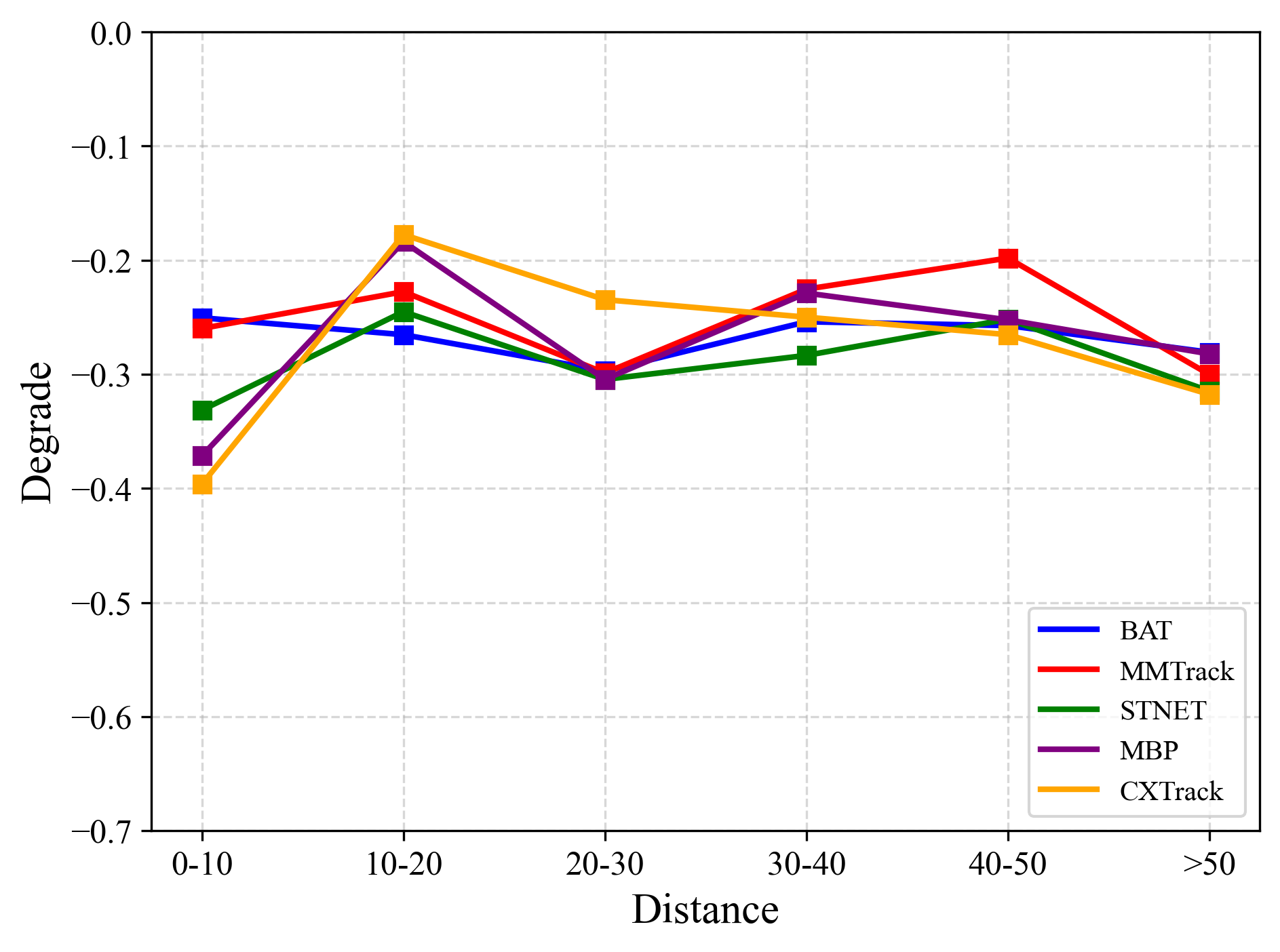}
    \end{minipage}%  
    \begin{minipage}[t]{0.32\linewidth}
    \centering
    \includegraphics[width=1\linewidth]{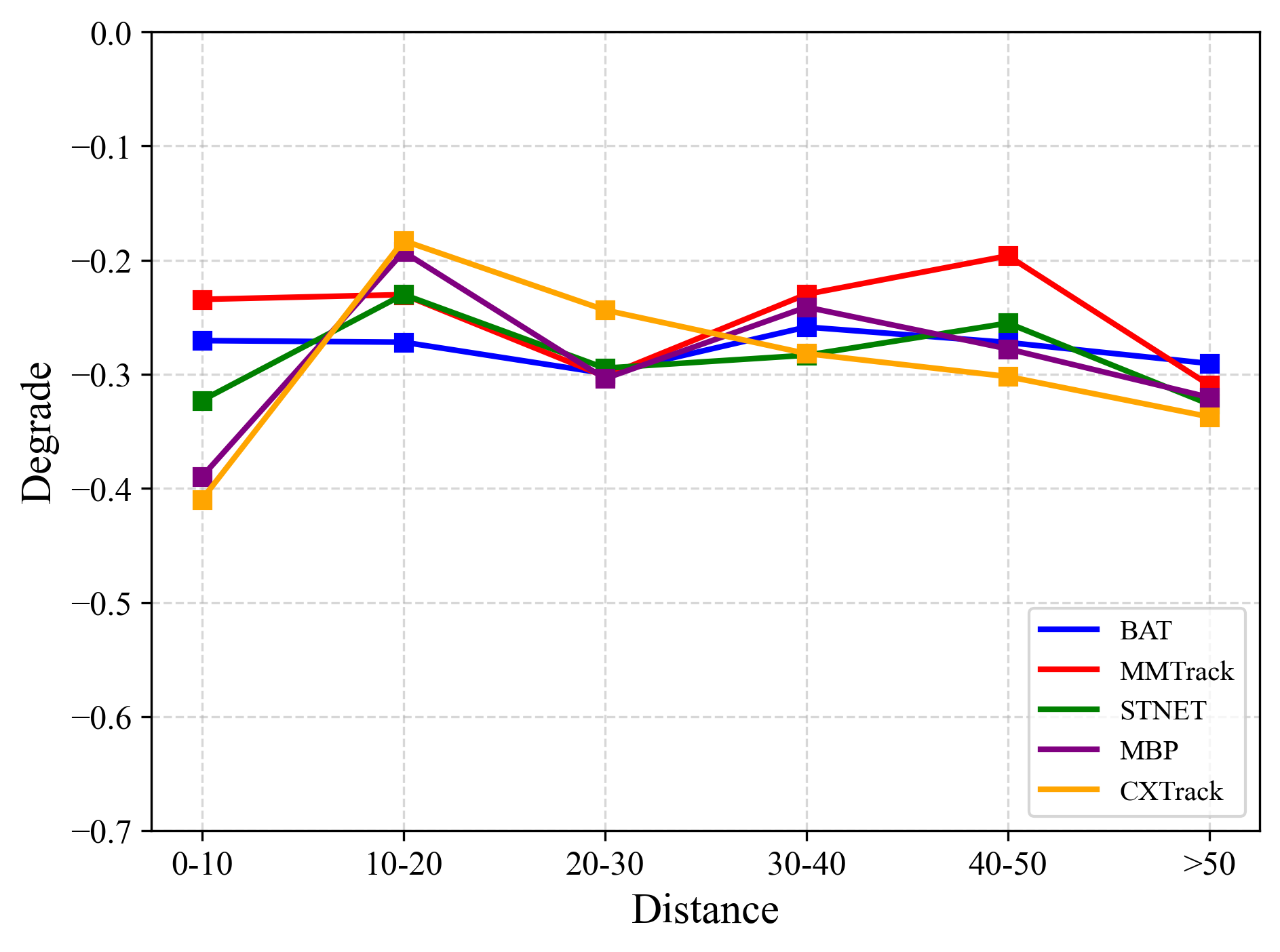}
    \end{minipage}% 
    \begin{minipage}[t]{0.32\linewidth}
    \centering
    \includegraphics[width=1\linewidth]{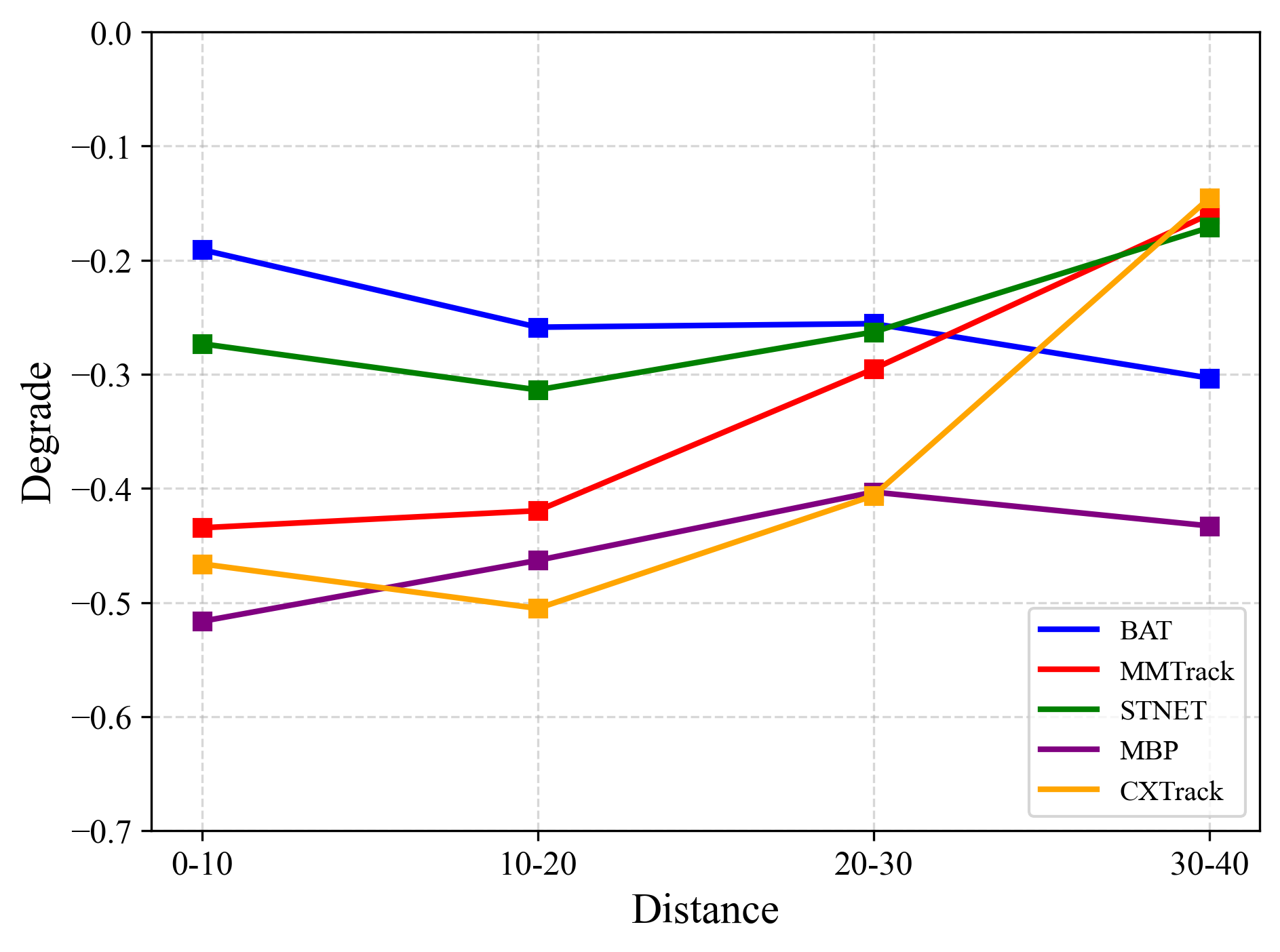}
    \end{minipage}%  
    \\
    \rotatebox{90}{~~~~~~~~Template Corruption}
    \begin{minipage}[t]{0.32\linewidth}
    \centering
    \includegraphics[width=1\linewidth]{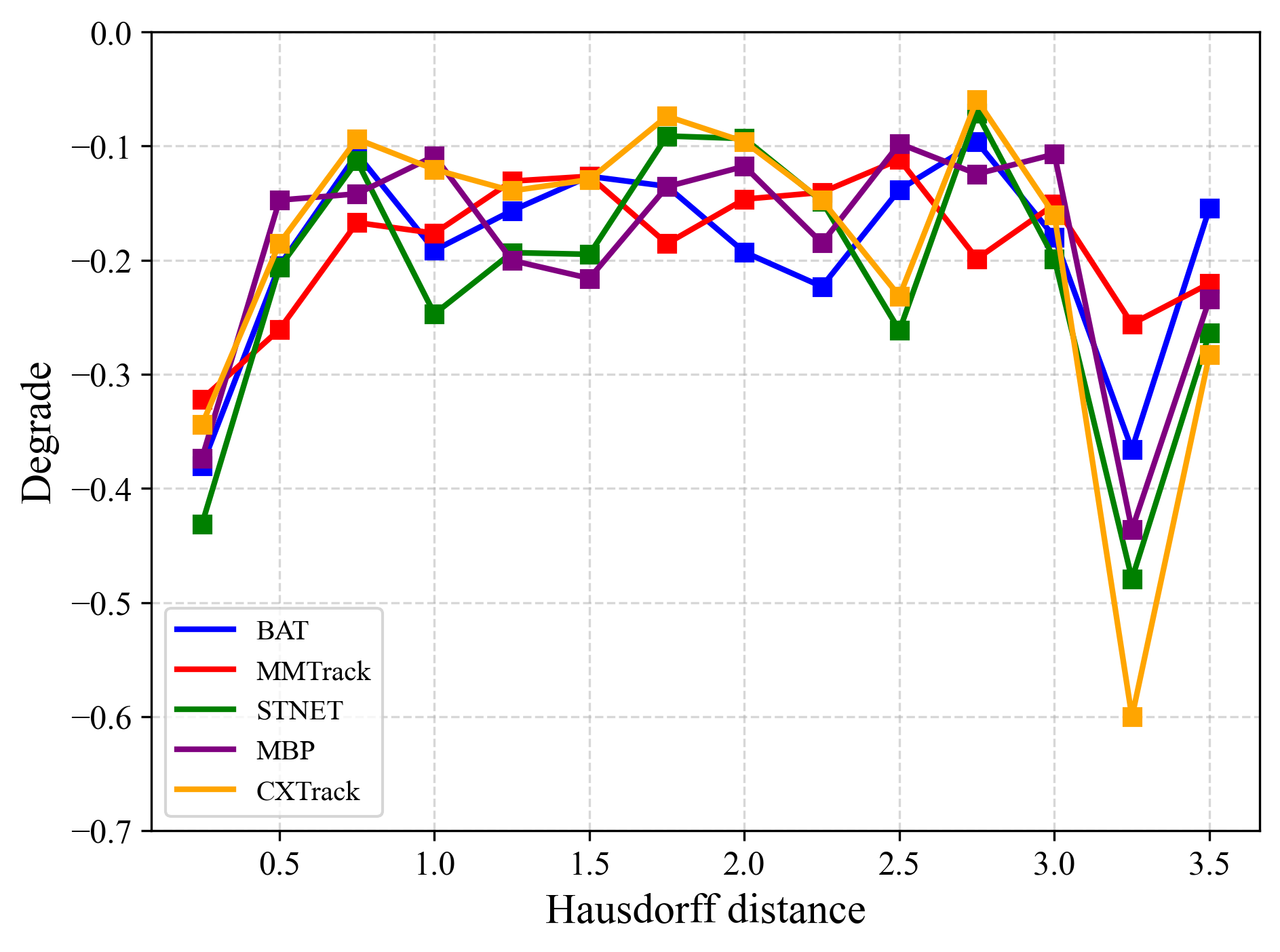}
    \end{minipage}%  
    \begin{minipage}[t]{0.32\linewidth}
    \centering
    \includegraphics[width=1\linewidth]{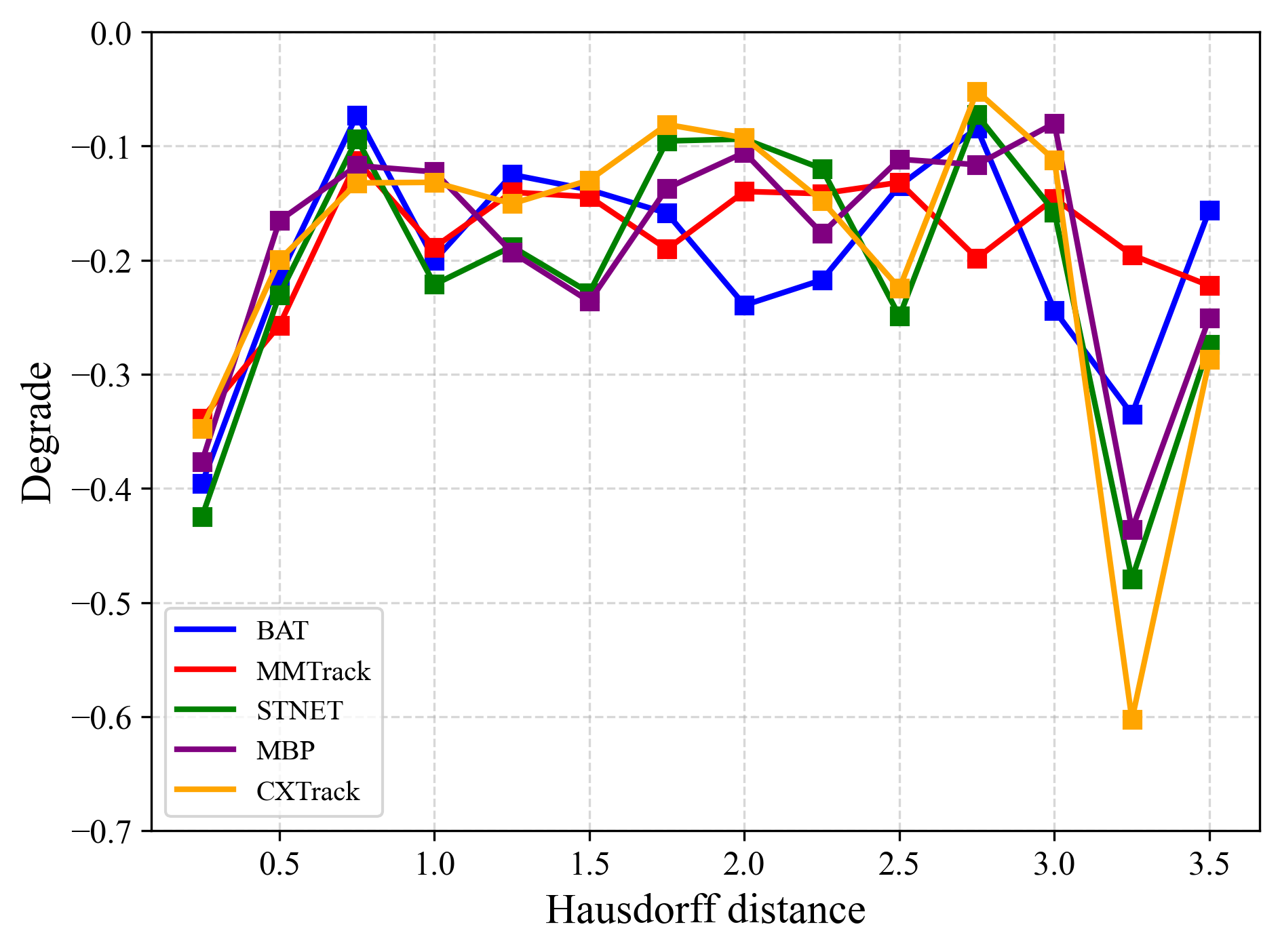}
    \end{minipage}% 
    \begin{minipage}[t]{0.32\linewidth}
    \centering
    \includegraphics[width=1\linewidth]{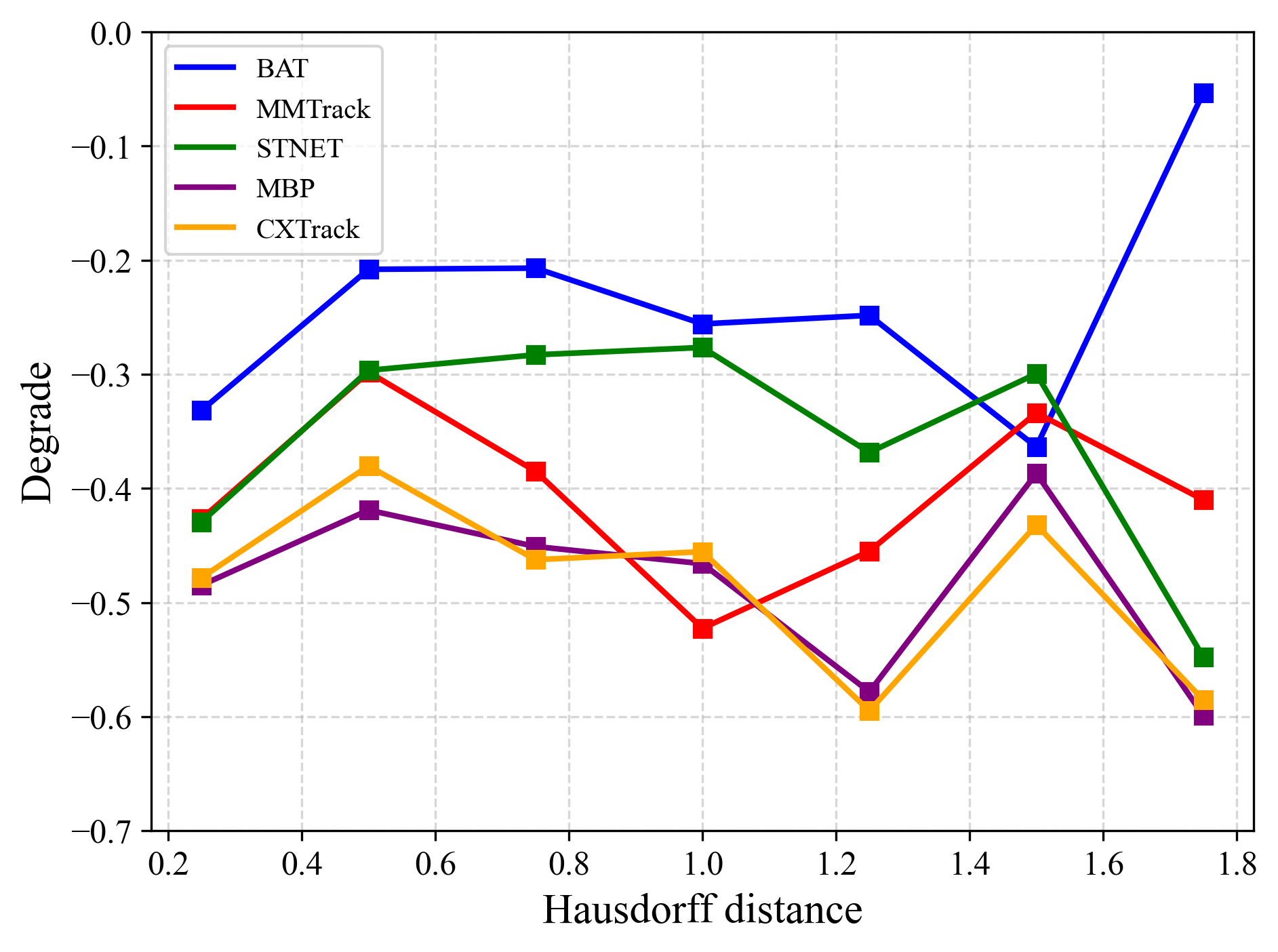}
    \end{minipage}% 
    \\
    \rotatebox{90}{~~~~~~~~~~Target Corruption}
    \begin{minipage}[t]{0.32\linewidth}
    \centering
    \includegraphics[width=1\linewidth]{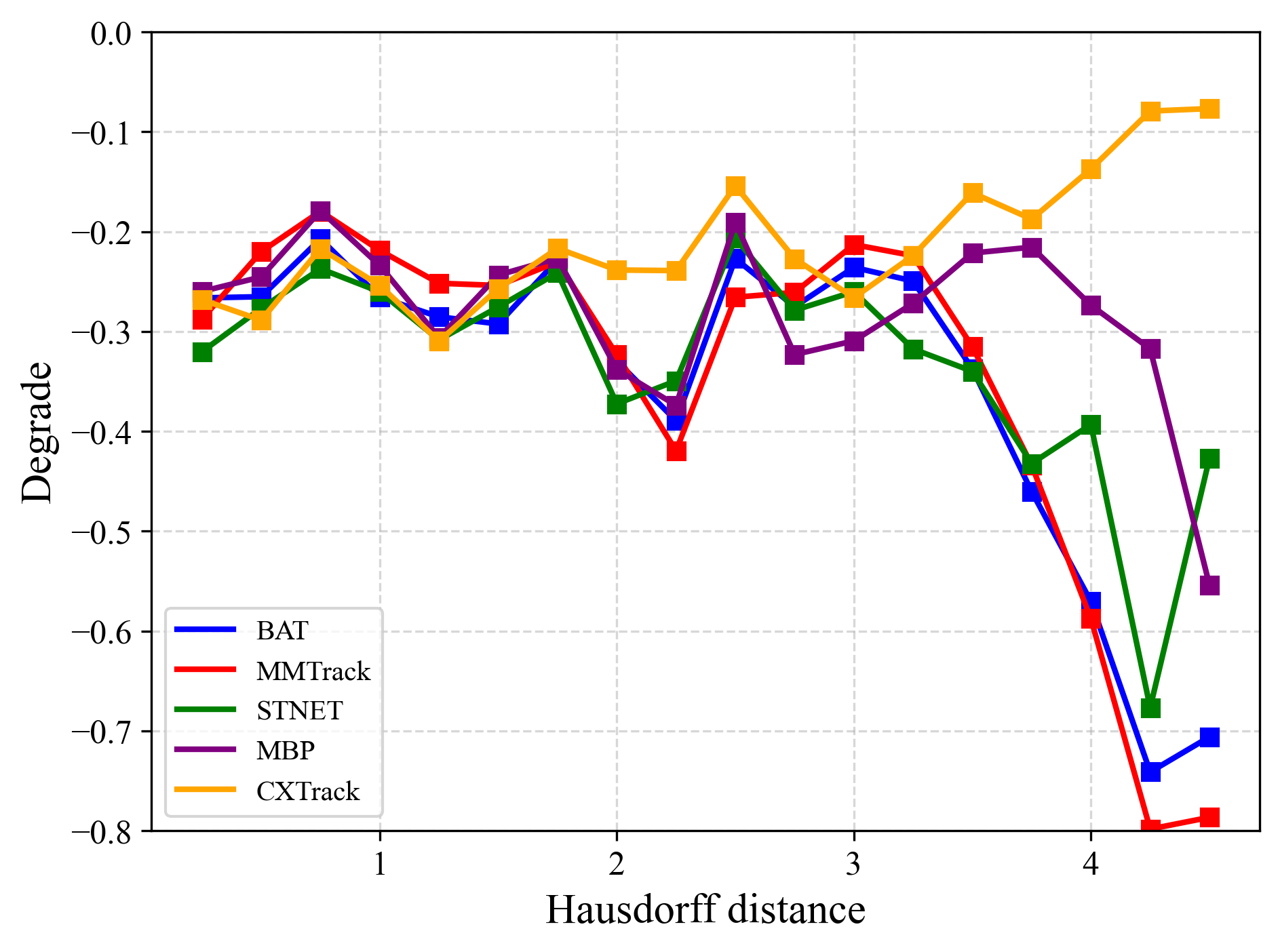}
    \end{minipage}%  
    \begin{minipage}[t]{0.32\linewidth}
    \centering
    \includegraphics[width=1\linewidth]{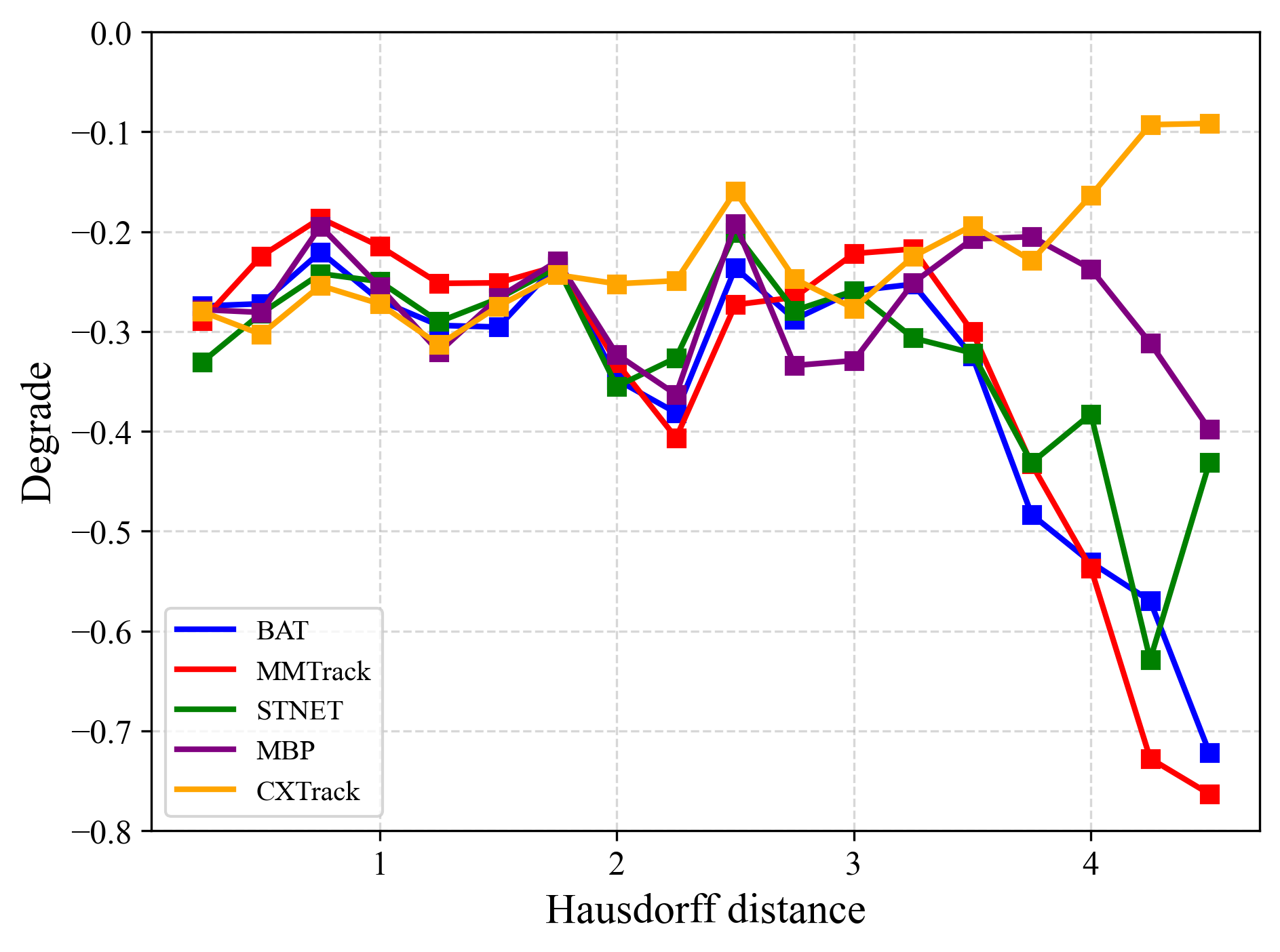}
    \end{minipage}% 
    \begin{minipage}[t]{0.32\linewidth}
    \centering
    \includegraphics[width=1\linewidth]{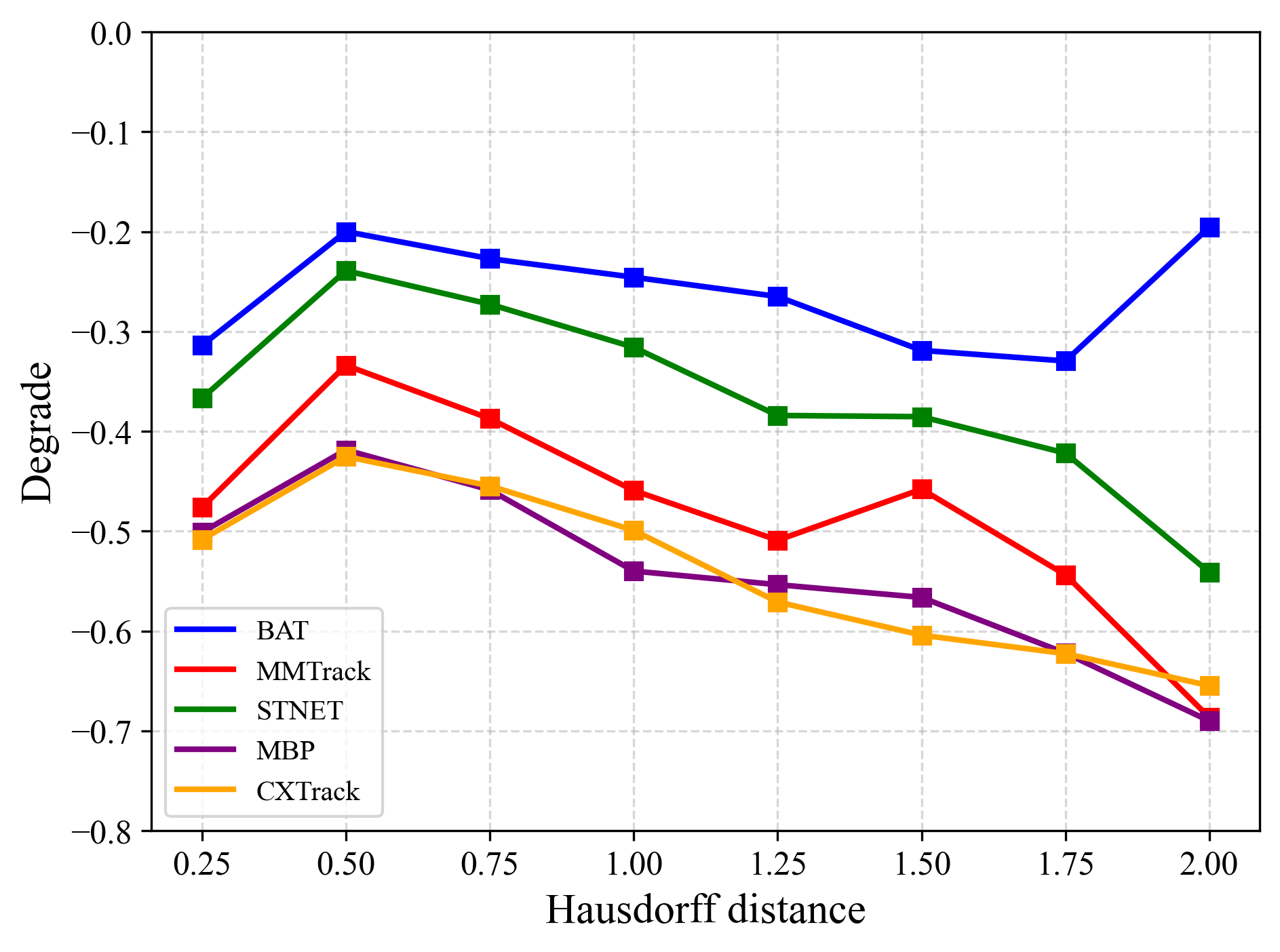}
    \end{minipage}% 
    \\
    \begin{minipage}[t]{0.33\linewidth}
    \centering
    (a) Rainy-Car
    \end{minipage}%
    \begin{minipage}[t]{0.33\linewidth}
    \centering
    (b) Snowy-Car
    \end{minipage}%  
    \begin{minipage}[t]{0.33\linewidth}
    \centering
    (c) Fogy-Pedestrian
    \end{minipage}%
    \caption{Analysis of the mean \textbf{\textit{IOU}} deviation in KITTI-A. The same column represents the same weather and category, and the same row represents the same factor. There are three factors: \textbf{\textit{Target Distance},} the horizontal axis represents the target distance; \textbf{\textit{Template Corruption},} the horizontal coordinates represent the Hausdorff distance between point clouds of templates in normal and adverse weather for the same tracking sequence; \textbf{\textit{Target Corruption},} the horizontal coordinates represent the Hausdorff distance between point clouds in normal and adverse weather for the same target. The vertical axis illustrates the IOU deviation between normal and adverse weather. 
    }
    \label{analyses}
\end{figure*}

\subsection{Robustness evaluations in synthetic data}\label{kitti-A_experimrnt}

In this subsection, we present the performance of current tracking methods on the synthetic datasets KITTI-A and nuScnenes-A. In the following paragraph, we will discuss the results and give insights separately according to weather type. Firstly, we examine the impact of adverse weather on targets at different distances. Secondly, we analyze the effect on tracking performance from the template perspective, considering the shape corruptions of the template due to adverse weather. Lastly, we investigate the impact on tracking performance from the perspective of the tracking object, considering the shape corruptions of targets. It's worth noting that we use the Hausdorff distance between clean point clouds and adverse point clouds to measure shape corruptions due to adverse weather.

\begin{table*}[htb!]
  \caption{The results of different trackers on nuScenes-A benchmark. The table documents the tracking success/precision of each method for various levels of adverse weather, along with the average degradation rate($DR$), range($R$), and standard deviation($S.d$). The best results are highlighted in bold.}
  \label{tab:nuScenes-results}
  \centering
  \scalebox{1.0}{
  \setlength{\tabcolsep}{1.0mm}{
  \begin{tabular}{c|c|c|ccccc|ccc}
    \toprule
      \multirow{2}{*}{Category} & \multirow{2}{*}{Method} & \multirow{2}{*}{clean} & \multicolumn{5}{c|}{nuScenes-A}  & \multirow{2}{*}{DR}  & \multirow{2}{*}{R}  & \multirow{2}{*}{S.d} \\
     &  &  & Lv-1 & Lv-2 & Lv-3 & Lv-4 & Lv-5 &  &  &  \\ 
    \midrule 
    \multicolumn{11}{c}{Rain}\\
    \midrule 
     \multirow{4}{*}{Car} 
    &BAT&41.126/43.653&24.899/24.836&25.007/24.923&24.949/24.858&25.015/24.959&25.102/24.982&0.39/0.43&0.20/0.15&0.08/0.06\\
    &MMTrack &57.211/65.726&27.397/34.957&27.272/34.802&27.315/34.806&27.222/34.777&27.222/34.762 &0.52/0.47&0.17/0.20&0.07/0.08\\
    &STNet &40.191/43.932&12.215/11.615 &12.215/11.608 &12.226/11.645 &12.214/11.588 &12.135/11.547&0.70/0.74&\textbf{0.09/0.10}&\textbf{0.04}/\textbf{0.04}\\
    &MBPTrack	&62.394/70.338&\textbf{38.687/43.183}&\textbf{38.654/43.156}&\textbf{38.718/43.266}&\textbf{38.750/43.168}&\textbf{38.665/43.124}& \textbf{0.38/0.39}&0.10/0.14&\textbf{0.04}/0.05\\
    \cmidrule{2-11}  		
     \multirow{4}{*}{Ped} 
    &BAT     &26.560/48.190 &12.075/19.419 &12.091/19.623 &12.097/19.539 &12.114/19.531 &12.158/19.754 & \textbf{0.54/0.59}& 0.08/0.34 &0.03/0.12  \\
    &MMTrack &34.646/60.174 &8.819/15.308 &8.822/15.294 &8.819/15.398 &8.814/15.285 &8.804/15.159 & 0.75/0.75 &0.02/0.24 &0.01/0.09  \\
    &STNet   &28.718/55.231 &9.179/17.554 &9.158/17.523 &9.164/17.499 &9.169/17.516 &9.166/17.421 & 0.68/0.68&	\textbf{0.02/0.13}& \textbf{0.01/0.05}\\
    &MBPTrack	 &45.352/74.064 &\textbf{14.991/24.107} &\textbf{14.988/24.089} &\textbf{15.021/24.147} &\textbf{15.098/24.267} &\textbf{15.075/24.318} &0.67/0.67&0.11/0.23 &0.05/0.10  \\
    \midrule 
    \multicolumn{11}{c}{Fog}\\
    \midrule 
     \multirow{4}{*}{Car} 
        &BAT & 41.126/43.653&26.578/26.855&26.518/26.793&26.621/26.981&26.883/27.298&29.458/30.435&\textbf{0.34/0.37} &2.94/3.64  &1.26/1.56  \\
        &MMTrack &57.211/65.726&27.593/35.121&27.399/34.975&27.593/35.201&27.672/35.365&27.367/35.344& 0.52/0.46 &  0.31/0.39 & 0.13/0.16 \\
        &STNet &40.191/43.932 &12.308/11.808&12.308/11.770&12.233/11.698&12.226/11.725&12.186/11.739& 0.70/0.73&\textbf{0.12/0.11} &\textbf{0.05/0.04}  \\
        &MBPTrack	&62.394/70.338&\textbf{39.152/42.859}&\textbf{39.115/42.901}&\textbf{38.820/42.691}&\textbf{38.743/42.599}&\textbf{39.761/43.578}& 0.37/0.39 & 1.02/0.98 &0.40/0.38  \\
      \cmidrule{2-11}  	
      \multirow{4}{*}{Ped} 
            &BAT &26.560/48.190&12.450/20.458&12.448/20.452&12.347/20.414&12.325/20.166&12.691/20.731& \textbf{0.53/0.58}&0.37/0.57  &0.14/0.20  \\
            &MMTrack &34.646/60.174&8.808/15.188&8.789/14.975&8.748/14.998&8.796/15.207&8.726/14.861& 0.75/0.75 & 0.08/0.35  & 0.03/0.15 \\
            &STNet &28.718/55.231 &9.212/17.771&9.202/17.741&9.212/17.719&9.238/17.748& 9.237/17.865& 0.68/0.68&\textbf{0.04/0.15}	&  \textbf{0.02/0.06}\\
            &MBP&45.352/74.064&\textbf{15.669/24.541}&\textbf{15.693/24.629}&\textbf{15.372/23.996}&\textbf{15.111/23.730}&\textbf{15.501/24.519}& 0.66/0.67 &0.58/0.90   &0.24/0.40  \\
    \midrule 
    \multicolumn{11}{c}{Snow}\\
    \midrule 
     \multirow{4}{*}{Car} 
    &BAT & 41.126/43.653 &24.888/24.748 &24.999/24.865 &25.201/25.199 &25.013/24.961 & 25.092/25.073 &0.39/0.43 &0.31/0.45  &0.12/0.18  \\
    &MMTrack &57.211/65.726 &27.373/34.975 &27.264/34.796 &27.066/34.544 &27.046/34.581 &27.054/34.564 & 0.53/0.47 &0.33/0.43   &0.15/0.19  \\
    &STNet &40.191/43.932 &12.156/11.560 &12.216/11.612 &12.157/11.561 &12.178/11.586 &12.101/11.500 & 0.70/0.74&\textbf{0.11/0.11}&\textbf{0.04/0.04}  \\
    &MBPTrack	&62.394/70.338 &\textbf{38.830/43.425} &\textbf{38.695/43.261} &\textbf{38.646/43.113} &\textbf{38.465/42.909} &\textbf{38.480/42.929} & \textbf{0.38/0.39} & 0.36/0.52  &0.15/0.22  \\
    \cmidrule{2-11}  		
    \multirow{4}{*}{Ped} &BAT &26.560/48.190 &12.077/19.441 &12.124/19.740 &12.218/19.874 &12.136/19.617 &12.066/19.468 & \textbf{0.54/0.59}&0.15/0.43  &0.06/0.18  \\
    &MMTrack &34.646/60.174 &8.836/15.408 &8.808/15.217 &8.842/15.269 &8.837/15.413 &8.871/15.502 & 0.74/0.74 &0.06/\textbf{0.29}   & \textbf{0.02/0.12} \\
    &STNet &28.718/55.231 &9.164/17.483 &9.166/17.625 &9.179/17.386 &9.163/17.404 &9.137/17.234 & 0.68/0.68&\textbf{0.04}/0.39	&\textbf{0.02}/0.14  \\
    &MBPTrack	&45.352/74.064 &\textbf{15.081/24.266} &\textbf{15.114/24.368} &\textbf{15.189/24.390} &\textbf{15.219/24.608} &\textbf{14.882/23.916} & 0.67/0.67 &0.34/0.69   &0.13/0.25  \\
    \bottomrule
   \end{tabular}
   }}
\end{table*}

\subsubsection{Robustness results on KITTI-A rainy}

We show the robustness results of 5 trackers on KITTI-A in Table~\ref{tab:KITTI-results}.
Generally, rainy days lead to a significant degradation in the cars, and pedestrians do not. However, different levels of rain exhibit minimal impact on its performance, and no discernible trend is observed.
Specifically, the performance of the methods in the cars decreases by 32\%-42\%, and MBPTrack~\cite{mbptrack} achieves the best performance with the lowest degradation, 34\% for success and 32\% for precision. However, when observing the differences between methods at different levels, MMTrack has the lowest standard deviation and range, which means that MMTrack is more stable than the other methods in rainy. In the pedestrian category, each method has a small performance drop except CXTrack \cite{cxtrack}, and the MMTrack is also stable. As depicted in Figure~\ref{analyses} (a), we analyze the results from three aspects: target distance, template shape corruption, and target shape corruption.

\textit{Analysis from target distance.}
Observing the degradation of performance for each method in targets at various distances, we note that rain has a more significant impact on closer cars, particularly in the range of 0-10 meters. This is attributed to LiDAR being more sensitive to rain at shorter distances. When comparing the degradation trends, BAT \cite{BAT} is relatively stable across different distances. This analysis suggests that box information may aid in maintaining stability across distances.

\textit{Analysis from template shape corruption.}
Observing the performance degradation of each method due to different template degradation, it becomes apparent that even minor changes in template shape (0-0.25) lead to significant performance degradation, with the IOU dropping by approximately 0.4. When the template shape degradation is in the range of 3.0-3.5, appearance-based methods such as BAT, STNet, MBPTrack, and CXTrack exhibit much higher IOU degradation compared to motion-based methods like MMTrack. This led us to investigate a motion-based approach to prevent degradation of the template shape.

\textit{Analysis from target shape corruption.}
Observing the performance degradation of each method under target degradation degrees, a fluctuating downward trend is evident as the target degradation increases. Particularly, when the Hausdorff distance exceeds 3, the performance of BAT, STNet, MMTrack, and MBPTrack decreases rapidly. In contrast, CXTrack exhibits more stable performance in the face of target degradation, CXTrack, This is made possible by the contextual information that CXTrack proposes to deal with the problem of large changes in appearance.

\subsubsection{Robustness results on KITTI-A snowy}

We show the robustness results of 5 trackers on KITTI-A in Table~\ref{tab:KITTI-results}. Specifically, the performance of the methods on the car category decreases by 34\%-43\%, with both MMTrack and MBPTrack decreasing by 34\%. And MMTrack has the lowest range($R$) and standard deviation($S.d$). As depicted in Figure~\ref{analyses} (b), we show an analysis of the car results from three aspects: target distance, template shape corruption, and target shape corruption. We can find each method has a similar performance in snow versus rain, which may mean snow and rain interfere with LiDARs in very similar ways. 

\subsubsection{Robustness results on KITTI-A fogy}
We show the robustness results of 5 trackers on KITTI-A in Table~\ref{tab:KITTI-results}. Under foggy weather conditions, the performance of methods on cars experienced a decrease ranging between 17\% and 24\%. Notably, the MMTrack method exhibited the best robustness with the least performance decrease at 18\%/17\%, showcasing a more stable performance with minimal changes across different fog levels. Conversely, foggy days had a more pronounced impact on pedestrians, resulting in a large performance degradation ranging from 51\% to 70\%, rendering the tracker ineffective in these conditions. As depicted in Figure~\ref{analyses} (c), we analyze the pedestrian results from three aspects: target distance, template shape corruption, and target shape corruption. 

\textit{Analysis from target distance.}
Observing the performance degradation of each method at different distances, as the distance increases, BAT performance recession increases, and MBPTrack decreases and then increases. One possible reason is that both methods rely on prior information from the bounding box. As the distance increases, foggy weather leads to missing points in the point cloud, impacting the association between the box and the points.

\textit{Analysis from template shape corruption.}
In terms of the degree of degradation of the template, it is observed that, to varying degrees, the performance degradation is in a state of fluctuation. There's nothing special about it.

\textit{Analysis from target shape corruption.}
As the degradation of the target increases, tracking performance for each method declines. This indicates that changes in the target's shape affect the tracker; as the shape degradation becomes more severe, there is a greater reduction in tracking performance.

\subsubsection{Robustness results on nuScenes-A}

We also evaluated the tracking method on the NuScenes-A dataset and discovered that adverse weather has a more significant impact on this dataset compared to KITTI-A. Specifically, as shown in Talbel \ref{tab:nuScenes-results}, we found the following: Firstly, in the cars, the performance of trackers shows greater degradation on nuScenes than on KITTI-A. Secondly, during rainy or snowy conditions, pedestrians experience more interference, which can even lead to complete tracking failures. 
From a sensor perspective, the point cloud data collected by NuScenes, using a 32-beam LiDAR, is sparser than the KITTI dataset, which employs a 64-beam LiDAR. This observation suggests that sparse data may be more susceptible to interference from adverse weather conditions.

\begin{table*}[htb]
    \caption{Performance comparison of advanced methods in CADC-SOT. Success/Precision at each level, mean values, range($R$) and standard deviation($S.d$) are reported. The best results are highlighted in bold.}
    \label{tab:CADC-result}
    \centering
    \begin{tabular}{l|l|llll|ll|lll}
    \toprule
    & & \multicolumn{4}{c|}{Snow}&\multicolumn{2}{c|}{Road}&~&~&~\\ 
    &Method & Light&Medium&Heavy&Extreme&Covered&Non covered & Mean & $R$ & $S.d$ \\
    \midrule
    \multirow{4}{*}{Car}
    &BAT &34.54/37.43 &30.22/32.55 &35.20/38.38 &28.41/29.73 &33.64/36.07 &31.70/34.24 &32.22/34.73 &6.79/8.65 &3.30/4.09\\ 
    &MMTrack &37.83/40.59 &40.53/43.91 &44.53/48.54 &39.71/42.34 &\textbf{42.14}/\textbf{45.46} &39.95/43.14 &40.54/43.76 &6.70/7.95 &2.82/3.41\\ 
    &MBPTrack &35.77/38.10 &\textbf{41.65}/\textbf{45.14} &\textbf{45.78}/\textbf{49.76} &\textbf{41.55}/\textbf{44.53} &41.10/44.45 &40.44/43.23 &40.92/44.12 &10.01/11.66 &4.11/4.80\\ 
    &STNet &\textbf{42.30}/\textbf{45.86} &40.08/43.55 &42.66/46.92 &38.79/42.11 &40.97/44.00 &\textbf{41.09}/\textbf{44.95} &\textbf{41.06/44.70} &\textbf{3.87}/\textbf{4.81} &\textbf{1.84}/\textbf{2.18}\\ 
    \cmidrule(lr){2-11}
    \multirow{4}{*}{Ped} 
    &BAT &23.77/29.61 &21.13/27.53 &42.07/58.86 &17.53/23.04 &27.64/35.33 &24.62/32.75 &25.33/33.35 &24.54/\textbf{35.83} &\textbf{10.93}/\textbf{16.30}\\ 
    &MMTrack &\textbf{36.92}/\textbf{47.87} &\textbf{35.09}/45.74 &58.17/78.69 &\textbf{24.50}/\textbf{31.14} &\textbf{42.17}/\textbf{55.38} &\textbf{37.70}/\textbf{49.49} &\textbf{38.74}/\textbf{50.86} &33.67/47.55 &14.10/19.99\\ 
    &MBPTrack &29.62/38.00 &34.74/\textbf{46.15} &\textbf{60.10}/\textbf{80.22} &22.76/29.63 &37.36/49.07 &35.21/46.05 &36.86/48.37 &37.34/50.59 &16.29/22.20\\ 
    &STNet &29.52/39.23 &23.68/31.71 &46.40/66.77 &22.65/30.01 &33.23/44.92 &28.22/38.60 &29.38/40.07 &\textbf{23.75}/36.77 &10.98/17.04\\ 
    \bottomrule
    \end{tabular}
\end{table*}

\subsection{Robustness evaluations in real-world data}

In this subsection, we present the performance of current tracking methods on the real-world dataset CADC-SOT. Since the CADC data only has snow weather and training on it doesn't fit our intent of evaluating robustness, we used parameters trained on nuScenes with the same LiDAR beams to test on CADC-SOT. Additionally, we further analyze the tracker's performance variation concerning target distance in snowy weather.

\begin{figure}[htb]
    \centering
    \begin{minipage}[t]{0.5\linewidth}
    \centering
    \includegraphics[width=1\linewidth]{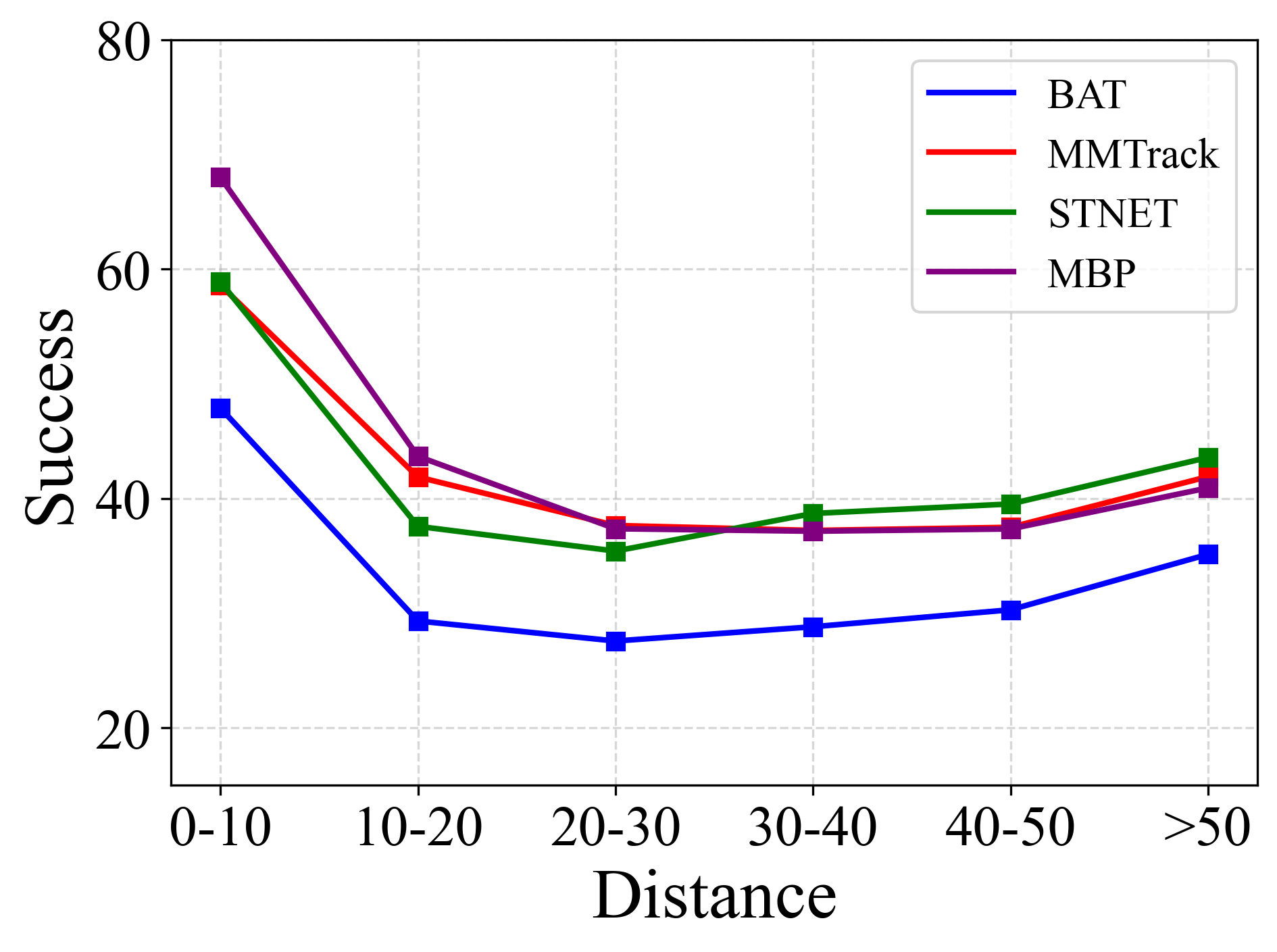}
    \end{minipage}%
        \begin{minipage}[t]{0.5\linewidth}
    \centering
    \includegraphics[width=1\linewidth]{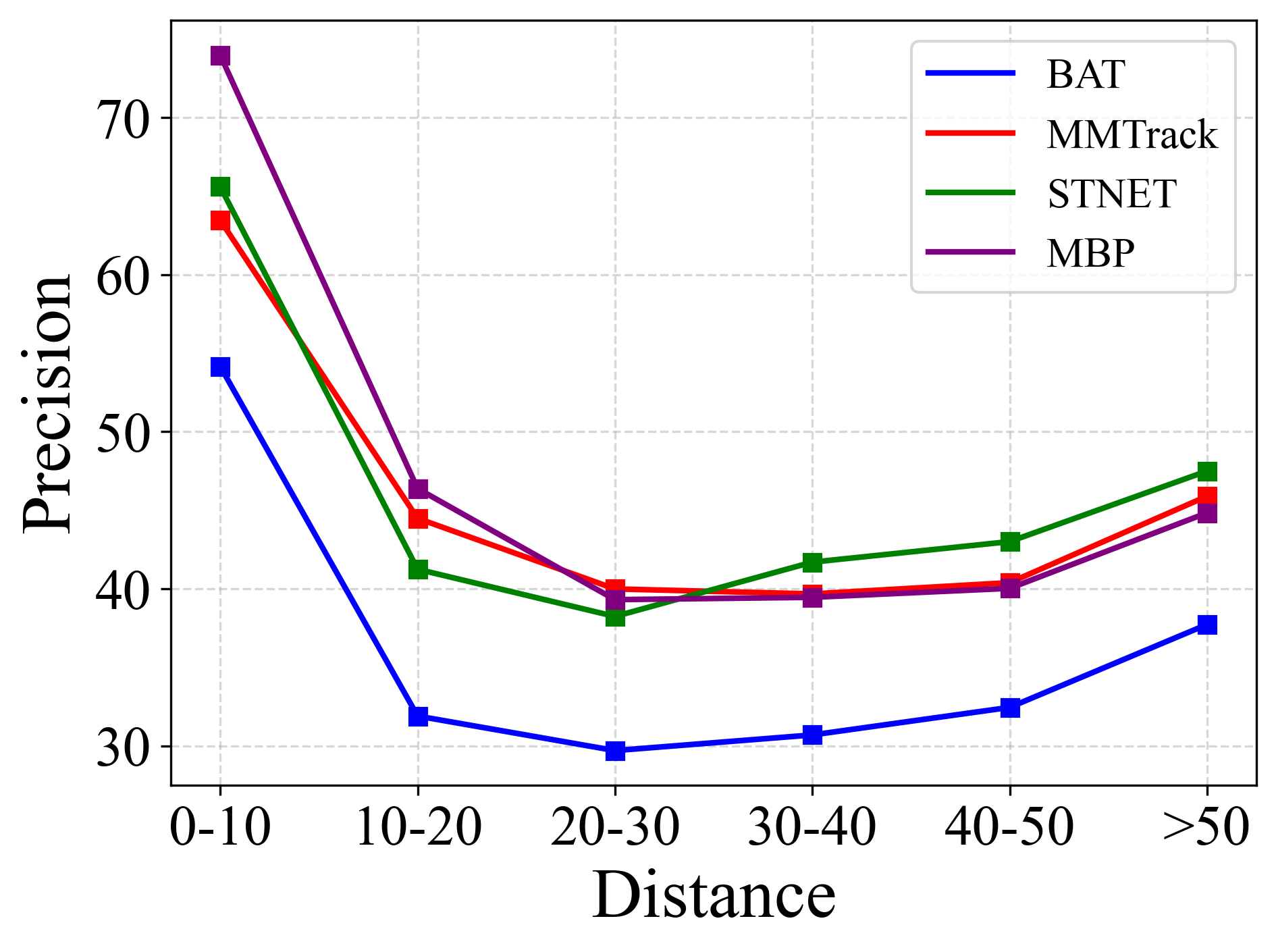}
    \end{minipage}%
    \\
    \begin{minipage}[t]{1.0\linewidth}
    \centering
    (a) Car
    \end{minipage}%
    \\
    \begin{minipage}[t]{0.5\linewidth}
    \centering
    \includegraphics[width=1\linewidth]{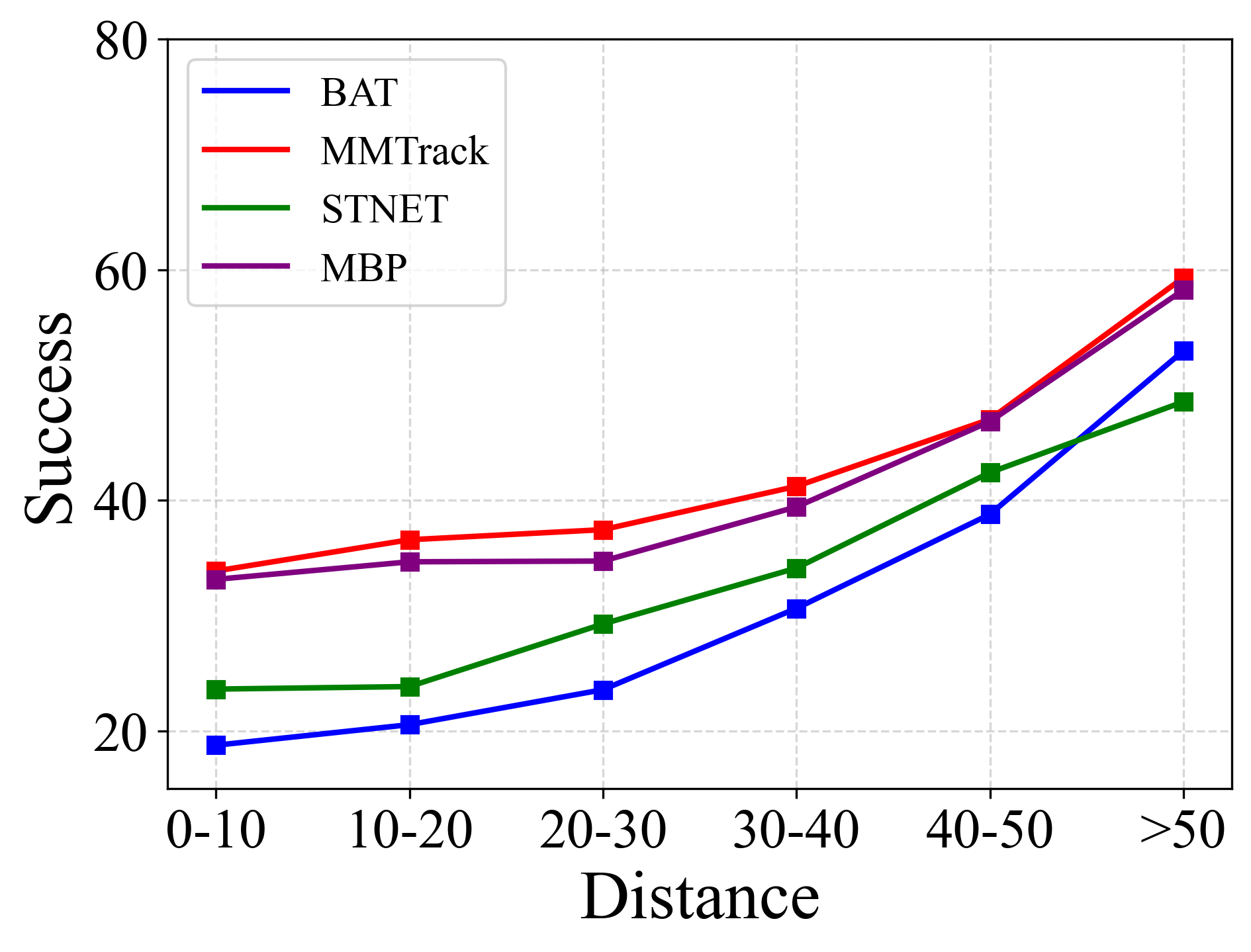}
    \end{minipage}% 
        \begin{minipage}[t]{0.5\linewidth}
    \centering
    \includegraphics[width=1\linewidth]{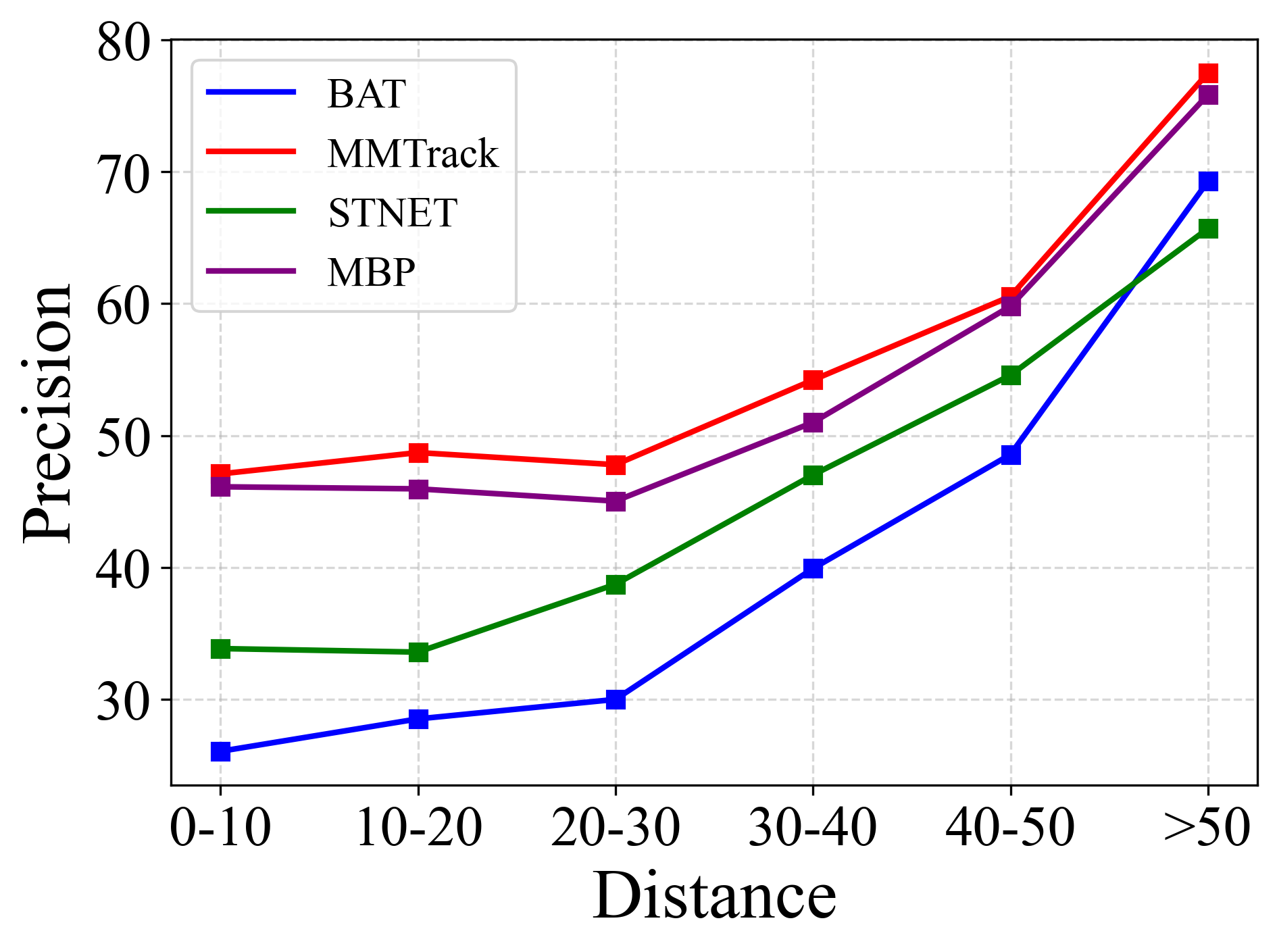}
    \end{minipage}% 
    \\
    \begin{minipage}[t]{1.0\linewidth}
    \centering
    (b) Pedestrian
    \end{minipage}%  
    \caption{The analysis in CADC-SOT, the horizontal axis represents the target distance, while the vertical axis illustrates the success or precision.}
    \label{fig:cadc_analyses}
\end{figure}

\subsubsection{Results}

As shown in Table~\ref{tab:CADC-result}, we present the performance of the current tracker on the real data CADC-SOT for different levels of snow and road conditions, respectively. We found that snow levels did not show a negative correlation with tracker performance under real weather conditions and that snow-covered roads did not lead to performance degradation.

In cars, STNet performed best in light snow, achieving success/precision of 42.30/45.86, and MBPTrack performed best in others, achieving success of 41.65/45.14, 45.78/49.74, and 41.55/44.53, respectively.
In the pedestrian category, MBPTrack performed best in heavy snow, achieving success/precision of 60.10/80.22, and MMTrack performed best in others, achieving mean success and precision of 38.74/50.86, respectively.

\subsubsection{Analysis}

The level of snowy days did not exhibit a clear correlation with performance. We further analyzed the effect of snowy days on the success of the tracker based on target distance. As shown in Figure.~\ref{fig:cadc_analyses} (a), snowy days have the most substantial impact on car targets in the 20-30 meter range and minimal impact on near targets. In Figure.~\ref{fig:cadc_analyses} (b), snowy days have the most significant effect on near targets in the pedestrian category, with better performance observed at longer distances. 

\section{Improving 3D Single Object Tracking Robustness}

\begin{figure*}[ht]
  \centering
  \begin{overpic}[width=2.0\columnwidth,tics=10]{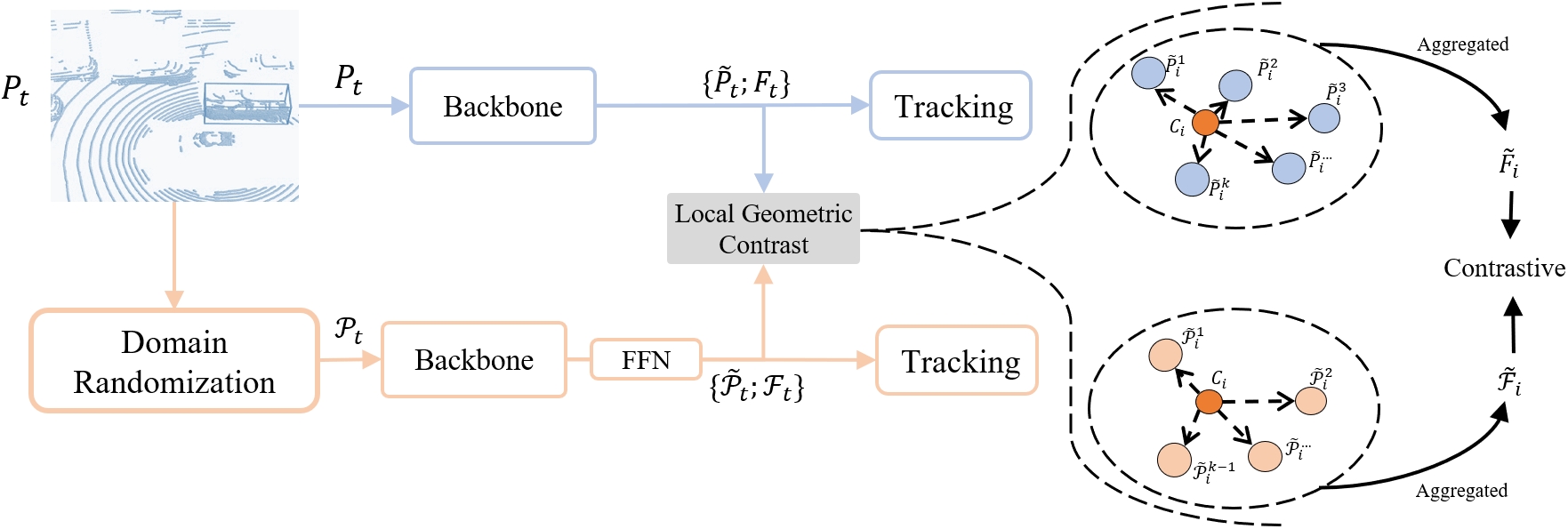}  
  \end{overpic}
  \caption{The overview of the proposed DRCT. DRCT consists of a primary branch (above) and an auxiliary branch (below), with a shared input of the point cloud $P_t$ at time $t$. The primary branch performs feature extraction on $P_t$ to obtain features $F_t$ and down-sampled points $\Tilde{P}_t$, while the auxiliary branch first generates a randomized point cloud  $\mathcal{P}_t$ through domain randomization, then extracts its features  $\mathcal{F}_t$ and down-sampled points $\Tilde{\mathcal{P}}_t$. The features from both branches are linked through local geometric contrastive learning, with $C_i$ being the common key points of both branches and neighborhood features are aggregated for contrastive learning.} 
  \label{fig:Network}
  \vspace{-7pt}
\end{figure*}

As can be seen from the above proposed benchmark, all advanced 3D trackers suffer from considerable degradation when dealing with adverse weather. In this section, we explored how to improve the ability of tacking such scenarios while maintaining performance in normal weather conditions. %  domain generalization 
By integrating domain randomization with contrastive learning, we designed DRCT, a method consisting of an auxiliary branch and a primary branch, that effectively enhances the robustness of models from normal to adverse weather environments. The approach is primarily composed of two modules: point cloud domain randomization and local geometric contrastive learning. Point cloud domain randomization strengthens the auxiliary branch's ability against corruption by generating various complex conditions; Local geometric contrastive learning enhances the robustness of the primary branch's feature representation in adverse weather scenarios by transferring the knowledge from the auxiliary branch, promoting the consistency of local geometric features. 

\subsection{Overview}

In contrastive learning techniques, the auxiliary branch typically possesses stronger performance or representational capacity in specific helpful domains. This study aimed to achieve the primary branch of DRCT to process complex scenarios, which require the auxiliary branch must have the ability to represent complex conditions, therefore, employing domain randomization techniques to create random domains for training the auxiliary branch. 
The primary branch is trained on clean data and transfers the representational capacity of the auxiliary branch through the local geometric contrastive learning module. 
This method not only ensures stability under clean weather but also enhances performance in adverse conditions. 

Figure \ref{fig:Network} illustrates the overall framework of our DRCT. Specifically, for the auxiliary branch, at time $t$, the input point cloud $P_t$ is randomly augmented through domain randomization techniques to generate the augmented point cloud $\mathcal{P}_t$. 
Subsequently, these augmented point cloud $\mathcal{P}_t$ are embedded through the backbone to obtain the point cloud features $\mathcal{F}_t$ and down-sampled points $\Tilde{\mathcal{P}}_t$.
For the primary branch, the input point cloud $P_t$ is directly embedded to obtain the features $F_t$ and $\Tilde{P}_t$. 
Then, we align the spatial positions of the two point clouds based on key points $C$, extract local geometric features, and perform contrastive learning. 
Finally, both branches continue to execute the subsequent tracking network, including relationship modeling and target position prediction. 

\begin{algorithm}
\caption{Domain randomization} 
\label{alg} 
\begin{algorithmic}[1]
    \renewcommand{\algorithmicrequire}{\textbf{Input:}}
	\renewcommand{\algorithmicensure}{\textbf{Output:}}
\REQUIRE Point cloud $P$ 
\ENSURE Augmented point cloud $\mathcal{P}$
    \STATE Initialize $\mathcal{P}=\emptyset$ 
    \STATE Initialize $ \mathcal{P}_{add\_noise}\gets \emptyset, P_{dropout} \gets \emptyset$
    \IF{Random()$<$0.5}
        \STATE \( \mathcal{P} \gets P \)\
    \ELSE
        \IF{Random() $>$ 0.5 }
            \STATE \( \text{Randomly generate the number of noise points } n \in (0, N_{max}) \)
            \STATE \( \text{Randomly generate Gauss noise points } \mathcal{P}_{add\_noise} \in R^{n \times 3} \)

        \ENDIF
        \\
        \IF{Random() $>$ 0.5}
            \STATE \( \text{Randomly generate the number of points to remove } r \)
            \STATE \( \text{Randomly generate index of remove points } I \in Z^{r}\)  
            \STATE \( P_{dropout} \gets P[I, :] \) 

        \ENDIF
        \\
        \IF{Random() $>$ 0.5}
            \STATE \( \delta _{\textit{jittering}} \gets \text{Generate jittering offsets } \delta \sim U(-a,a) \)
        \ELSE
            \STATE \(  \delta _{\textit{jittering}} \gets \mathbf{0} \)
        \ENDIF

        \STATE \( \mathcal{P} = \left( P \cup \mathcal{P}_{\textit{add\_noise}} \setminus P_{\textit{dropout}} \right) + \delta _{\textit{jittering}} \)
    \ENDIF
    \RETURN \( \mathcal{P} \)
\end{algorithmic}
\end{algorithm}

\subsection{Domain Randomization}

Domain randomization technique \cite{semaniticstf, DR} is an effective way to enhance model robustness and generalization ability, as randomized augmentation can create complex and diverse samples, thereby improving the model's representational capacity for these complex samples.
In our DRCT, we use domain randomization techniques to shift the distribution of the input point clouds; specifically, we employ three augmentation steps: adding noise, dropout points, and jittering points.
For the input point cloud $P$, the point cloud augmentation can be expressed in the following form:

\begin{equation}
\begin{aligned}
\mathcal{P} = \left( P \cup \mathcal{P}_{\textit{add\_noise}} \setminus P_{\textit{dropout}} \right) + \delta _{\textit{jittering}}
\end{aligned}
\end{equation}

where $ \mathcal{P}_{\textit{add\_noise}} $ means the noise set; $ \mathcal{P}_{\textit{dropout}} $ means the remove points set; and the $ \delta _{\textit{jittering}} $ means jittering offsets. The complete process is shown in Algo \ref{alg}.

\begin{table*}[htb!]
 \centering
  \caption{Performance comparison with the baseline method on the KITTI-A benchmark. We document the success/precision for various types of adverse weather in the table. The performance improvements are shown in the last row.}
  \label{tab:drct_kitti_result}
  \scalebox{1.0}{
  \setlength{\tabcolsep}{1.0mm}{
  \begin{tabular}{c|c|c|ccccc|c}
    \toprule
       \multirow{2}{*}{~} & \multirow{2}{*}{Method} & \multirow{2}{*}{Clean} & \multicolumn{6}{c}{KITTI-A} \\
      & &  & Lv-1 & Lv-2 & Lv-3 & Lv-4 & Lv-5 &  Mean \\ 
    \midrule  
       \multicolumn{9}{c}{Rain}\\
    \midrule
        \multirow{2}{*}{Car} 
         & MBPTrack & 73.70/85.22 & 49.28/58.94 & 48.32/57.58 & 46.21/55.11 & 51.11/61.36 & 48.64/57.83 & 48.71/58.16 \\
         & DRCT & 75.30/86.33 & 50.87/60.32 & 52.69/62.60 & 49.92/59.15 & 51.95/61.47 & 50.69/59.92 & 51.22/60.69 \\ 
         %\cmidrule(lr){3-9}
         \rowcolor{mypink} &      & +1.60/+1.11 & +1.59/+1.38 & +4.37/+5.02 & 	+3.71/+4.04 & +0.84/+0.11 & +2.05/+2.09 & +2.51/+2.53    \\ 
    \cmidrule(lr){2-9}
        \multirow{2}{*}{Ped} 
         & MBPTrack & 66.00/90.92 & 65.26/89.49 & 65.4/90.89 & 65.91/90.88 & 64.31/88.64 & 67.40/92.46 &  65.66/90.47 \\
         & DRCT &  67.75/92.90 & 66.00/90.20 & 64.60/87.66 & 63.47/87.01 & 66.22/89.97 & 67.81/92.83  &  65.62/89.53 \\
            % 67.30/92.40  & 
          %\cmidrule(lr){3-9}
         \rowcolor{mypink} & & +1.75/+1.92&  +0.74/+0.71 & -0.8/-3.23 & -2.44/-3.87 &+1.91/+1.33  &  +0.41/+0.37 & -0.04/-0.94 \\
        \midrule 
       \multicolumn{9}{c}{Fog}\\
        \midrule 
        \multirow{2}{*}{Car} 
         & MBPTrack & 73.70/85.22 & 59.28/70.13 & 56.70/66.98 & 56.39/66.64 & 56.41/66.71 & 56.32/66.56& 57.02/67.40 \\
         & DRCT & 75.30/86.33 & 59.77/70.06 & 58.03/67.95 & 57.88/67.94 & 57.81/67.86 & 57.74/67.66  & 58.25/68.29   \\ 
         % \cmidrule(lr){3-9}
        \rowcolor{mypink}  &  & +1.60/+1.11 & +0.49/-0.07 & +1.33/+0.97 & +1.49/+1.30 & +1.40/+1.15 & +1.42/+1.10 & +1.23/+0.89   \\ 
    \cmidrule(lr){2-9}
        \multirow{2}{*}{Ped} 
         & MBPTrack             & 66.00/90.92  & 33.63/54.12 & 25.30/42.66 & 13.95/27.68 & 14.46/27.72 & 14.09/27.58 &  20.29/35.95\\
         % & DRCT & 67.34/92.12 & 29.796/48.988 &  23.403/40.262 & 13.264/25.821 & 13.219/25.928 & 13.885/26.663 &  \\
         & DRCT                 & 67.75/92.90  & 34.60/57.40 & 27.32/47.97 & 15.13/28.88 & 14.97/28.46 & 14.62/28.05 & 21.33/38.15 \\
          %\cmidrule(lr){3-9}
         \rowcolor{mypink}  &   & +1.75/+1.92  & +0.97/+3.28 & +2.02/+5.31 & +1.18/+1.2  & +0.51/+0.74 & +0.53/+0.47 &  +1.04/+2.2 \\
       \midrule
       \multicolumn{9}{c}{Snow}\\
       \midrule 
       \multirow{2}{*}{Car}  
         &MBPTrack & 73.70/85.22 & 49.19/58.71 & 48.11/57.06 & 48.27/57.45 & 45.74/54.27 & 46.09/54.94 &  47.48/56.49 \\
         & DRCT & 75.30/86.33 & 51.21/60.61 & 51.24/60.60 & 51.27/60.80 & 47.09/55.36 & 46.01/53.80 & 49.36/58.23   \\ 
         % \cmidrule(lr){3-9}
         \rowcolor{mypink} & &+1.60 / +1.11 & +2.02/+1.90 & +3.13/+3.54 & +3.00/+3.35 & +1.35/+1.09 & -0.08/-1.14 & +1.88/+1.74  \\ 
    \cmidrule(lr){2-9}
        \multirow{2}{*}{Ped} 
         & MBPTrack          & 66.00/90.92  & 66.78/91.53    & 65.40/90.22 & 65.34/90.20 & 64.87/89.29 & 64.72/88.60 & 65.42/89.97 \\
         & DRCT              & 67.380/92.71 & 65.14/88.84    & 66.45/90.92 & 67.25/91.98 & 65.38/89.18 & 65.77/89.44 & 66.00/90.07  \\
         % \cmidrule(lr){3-9}
        \rowcolor{mypink} &  & +1.75/+1.92  & -1.64/-2.69    & +1.05/+0.7  & +1.91/+1.78 & +0.51/-0.11 & +1.05/+0.84 & +0.58/+0.1 \\
    \bottomrule
  \end{tabular}
  }}
\end{table*}

\subsection{Local Geometric Contrast Learning}

Most contrastive learning methods focus on spatially aligned representations, typically using BEV-features or voxel-features for contrastive learning.
However, for tracking tasks, point-based backbone networks perform better. In this work, we consider using point-based contrastive learning, but the lack of one-to-one correspondence between the point clouds of the two branches complicates feature alignment.
Therefore, we propose a local geometric contrastive learning module to address this issue, which can align the neighborhood features of the two branch point clouds.
Specifically, inputting the normal point cloud $P_t$, the primary branch processes $P_t$ to obtain the features $F_t$ after feature embedding through the backbone, along with the down-sampled point clouds$\Tilde{P}_t$. In the auxiliary branch, a feedforward neural network (FFN) module is utilized to further embed features followed the backbone, by inputting the augmented point cloud $\mathcal{P}_t$, resulting in the auxiliary branch features $\mathcal{F}_t$ and down-sampled points $\Tilde{\mathcal{P}}_t$.
Then, treating the points in $\Tilde{\mathcal{P}}_t$ as key points $C$ for local geometric contrastive learning, denoting $C_{i} \in \Tilde{\mathcal{P}}_t$ as the i-th to-be-contrasted point, we can find its K neighboring $ \{ \mathcal{N} _{C_i}  \mid \Tilde{\mathcal{P}}_i^{1},\Tilde{\mathcal{P}}_i^{2},...,\Tilde{\mathcal{P}}_i^{k} \} \in R^{k \times 3}$ in $\Tilde{\mathcal{P}}_t$ by Ball Query, and aggregate local geometric feature $\Tilde{\mathcal{F}}_{i}$ for it by Max Pooling.
This process can be formulated as:
\begin{equation}
\begin{aligned}
\Tilde{\mathcal{F}}_{i} = MaxPooling(Ballquery(C_{i}, \Tilde{\mathcal{P}}_t,K))
\end{aligned}
\end{equation}

Next, we extract $K$ nearest neighbor features centered on $C$ under point cloud $\Tilde{P}_t$ on the primary branch, and features $\Tilde{F}_t$ are obtained to contrast.
Finally, the training objective of local geometric contrastive learning can be formulated as

\begin{equation}
\begin{aligned}
\arg\min \mathbb{E} \left[ \frac{1}{N} \sum_{i=1}^{N} \left\lVert \Tilde{\mathcal{F}}_{i} - \Tilde{F}_{i} \right\rVert \right]
\end{aligned}
\end{equation}

In this module, we align the point clouds of the two branches through common key points, using local neighborhood-based geometric contrastive learning to constrain the consistency of features between the two branches. Based on this, the primary branch can effectively understand feature variations under adverse conditions. Furthermore, as this branch is trained on clean point clouds, it also ensures stability under normal weather conditions. 

\subsection{Experimental Verification}

To evaluate our method's performance in adverse weather and verify its robustness enhancement capability, we performed experiments on benchmark datasets. These experiments consist of comparisons with baseline methods and ablation studies, providing a thorough evaluation of the effectiveness and advantages of our approach.
We adopt MBPTrack \cite{mbptrack} as our baseline method, a point-based tracker that utilizes DGCNN \cite{dgcnn} for feature extraction and employs a memory mechanism to retain information from previous frames, modeling the relationship between the current and previous frames based on this memory and finally using a 3D CNN as the task head to produce tracking results. In our approach, we apply an MBPTrack network as the primary branch, then we duplicate one to serve as the auxiliary branch and append a feedforward neural network following its backbone to enhance feature extraction. Information is transferred between the two branches through the local geometric contrastive learning module, thereby enhancing the model's robustness and performance.

\begin{figure*}[htb]
  \centering
  \begin{overpic}[width=2.0\columnwidth,tics=10]{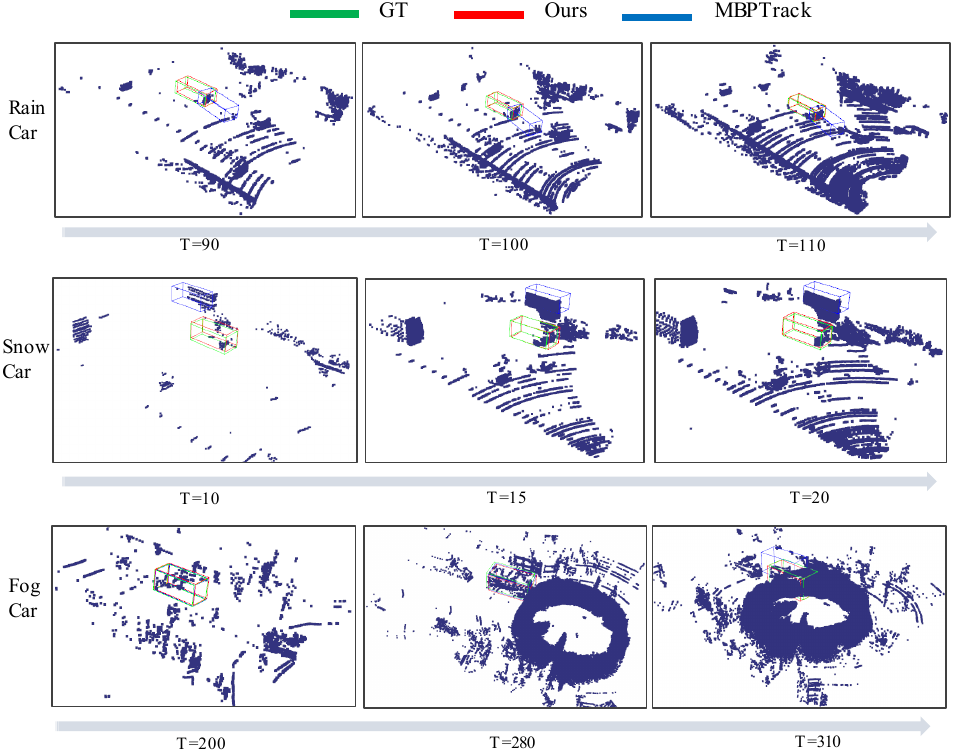}
  \end{overpic}
  \caption{Visualization results demonstrate that we proposed method is competitive in adverse weather. The rows from top to bottom indicate rain, snow, and fog, respectively. And the ground truth is plotted in green, and the result boxes are plotted in blue and red for MBPTrack and our method, respectively.}
  \label{fig_results:adverse_vis}
  \vspace{-7pt}
\end{figure*}

We evaluated the adverse weather performance of our proposed DRCT on the KITTI-A. In table \ref{tab:drct_kitti_result}, it is observed that our method improves the performance of the baseline method in almost all conditions. For the car category, with obvious improvements, our method outperforms MBPTrack by +2.51/+2.53, +1.23/+0.89, and +1.88/+1.74 mean success/precision for rain, fog, and snow, respectively. These improvements demonstrate the positive effects of our method for MBPTrack, which benefits from contrastive learning that can successfully transfer assist knowledge to the primary branch.
For the pedestrian category, our method delivers better results in most weather conditions, particularly in foggy conditions. 
Furthermore, Fig. \ref{fig_results:adverse_vis} visualizes the tracker results of the proposed DRCT and MBPTrack for comparison. Our method more accurately localizes those objects in adverse weather conditions with the target missing for MBPTrack. 

\begin{table}[htb!]
\renewcommand{\arraystretch}{1.1}
 \centering
  \caption{The comparison with different strategies of training. We document the mean success/precision of each method for various adverse weather on KITTI-A. The performance improvements are shown in the last row.}
  \label{tab:ablation study ra}
  \scalebox{0.8}{
  \setlength{\tabcolsep}{1.0mm}{
  \begin{tabular}{c|ccccc}
    \toprule
     \multirow{2}{*}{Method} &  \multicolumn{5}{c}{KITTI} \\
       &  Clean & RD & Rain & Fog & Snow  \\ 
    \midrule  
        % \multirow{4}{*}{Car} 
                    MBPTrack      & 73.70/85.22 & 41.32/47.64 & 48.71/58.16  & 57.02/67.40 &  47.48/56.49   \\
                    MBPTrack-RD   & 69.77/80.94 & 65.10/75.48 & 46.34/56.54  & 56.94/67.43 &  44.55/54.10   \\
        % \cmidrule(lr){2-6}
                    DRCT-Auxiliary   & 62.72/71.13 & 62.50/72.34 & 41.29/46.38  & 61.81/72.55 &  40.87/45.93   \\
                    \cmidrule(lr){2-6}
         
            \multirow{1}{*}{DRCT} & 75.30/86.33 & 56.60/64.95 & 51.22/60.69  & 58.25/68.29 &  49.36/58.23   \\
                \rowcolor{mypink} & +1.6/+1.11   & +15.28/+17.31 & +2.51/+2.53  & +1.23/+0.89 & +1.88/+1.74    \\
    % \cmidrule(lr){1-7}
    %     \multirow{3}{*}{Ped} 
    %      & MBPTtack & 68.6/93.9 & & 65.66/90.47 & 20.29/35.95& 65.42/89.97   \\
    %      & MBP-RA & 73.70 & \textbf{49.28} & \textbf{48.32} & 46.21 & \textbf{51.11} \\
    %      & DRCT & 65.21 & 54.65 & 60.05 & 58.49 & 58.58  \\
    \bottomrule
  \end{tabular}
  }}
\end{table}

\subsection{Ablation study}

To further validate the effectiveness and reasonableness of our DRCT, we designed ablation studies from two aspects: training strategy and contrastive learning architecture, beginning with verifying the effectiveness of the training strategy, and followed by an evaluation of the contrastive learning structures.

\subsubsection{Compare with different training strategies}

We developed a local geometric contrastive learning module based on contrastive learning to facilitate knowledge transfer from the auxiliary branch to the primary branch during joint training. To evaluate the effectiveness of this training strategy, we compared it to one strategy that uses the original MBPTrack framework and directly feeds the samples augmented by random domain. As shown in Table \ref{tab:ablation study ra}, the method that trains directly in random domains, called MBPTrack-RD, achieves competitive performance in random domains but is unsatisfactory in adverse weather conditions. In contrast, our strategy enables the primary branch to inherit the merit of the auxiliary branch in random domains, being capable of handling complex conditions. As a result, our method shows superior performance in adverse weather conditions, with improvements of +15.28/+17.31 in random areas, +2.51/+2.53 in rain, +1.23/+0.89 in fog, and +1.88/+1.74 in snow.

\begin{figure}[htb!]
  \centering
  \begin{overpic}[width=1.0\columnwidth,tics=10]{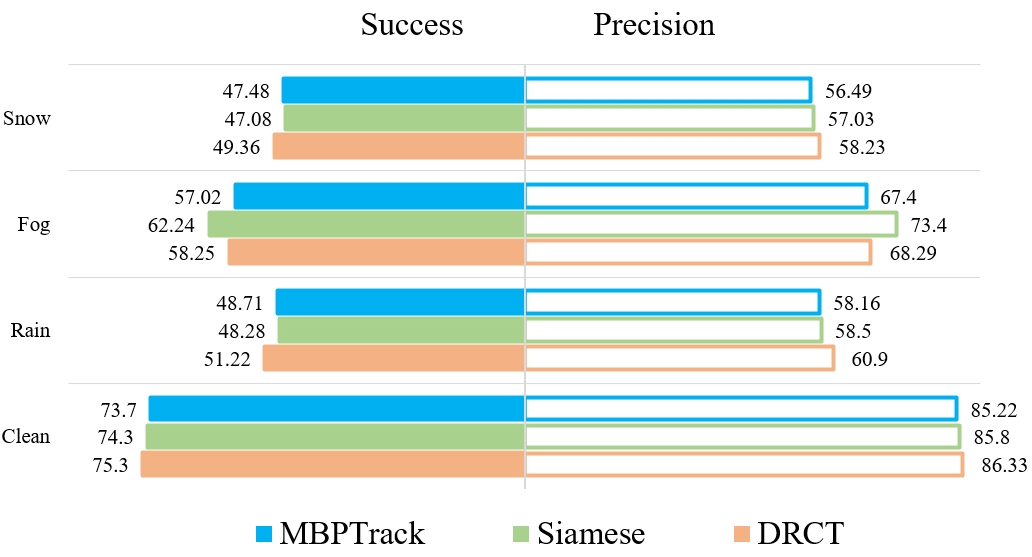}
  \end{overpic}
  \caption{The comparison with different contrastive learning structures. In this figure, the structure two branches have the shared parameters named Siamese, colored as green, while our DRCT is colored orange and the baseline method is blue. The tracking metric success and precision are plotted on it, and success is on the left side and precision on the right.}
  \label{fig:ablation}
  \vspace{-7pt}
\end{figure}

\subsubsection{The effect of different architectures}
 
In fact, there are two different architectures to conduct the local geometry contrastive learning. The traditional contrastive learning adopts the Siamese network that shares the parameters of each component between the primary and auxiliary branches. Different from it, considering that each branch should have knowledge of a different domain, our DRCT uses an asymmetric dual-branch architecture, where the auxiliary branch has extra FFN than the primary branch. 

Figure \ref{fig:ablation} shows the comparison results on cars of KITTI-A dataset. As can be seen, compared to the original MBPTrack, the Siamese architecture shows significant improvement in foggy conditions, but gets slightly degraded in rainy and snowy weather. 
The explanation may lie in the fact that the random domain requires more robust and varied perturbations to cover the complex distribution of point clouds in foggy conditions. However, these random augmentations can lead to overfitting in the Siamese network. As a result, the network may become overly specialized in adapting to shape changes under fog, making it struggle to adjust to rainy and snowy environments.
In contrast, our method demonstrates improvements across all weather conditions, highlighting its stronger generalization capability. This aligns with our original design intent to transfer the auxiliary branch's representational ability to process complex conditions rather than overfitting to the random domain.

\section{Conclusion}

In this paper, we design a benchmark for adverse weather that is suitable for 3D single object tracking robustness evaluation. This benchmark contains both synthetic and real-world data for three types of weather: rainy, foggy, and snowy, each with five levels. After our robustness evaluation of the state-of-the-art methods, we further analyze each tracker on three aspects: object distance, template corruption, and target corruption. To defend against degradation under adverse weather, based on the domain randomization technique, we proposed a local geometric contrastive learning module, effectively improving the robustness of the tracker. We hope that our benchmarks, evaluations, and analysis will inspire future 3D SOT tasks in adverse weather.

\section*{Acknowledgment}
This work is supported in part by the National Natural Science Foundation of China (Grand No. 62272082, No. 62301562), the China Postdoctoral Science Foundation (Grand No. 2023M733756), and the Fundamental Research Funds for the Central Universities (Grand No. 2023QN1055).

%\ifCLASSOPTIONcaptionsoff
%  \newpage
%\fi

\bibliographystyle{IEEEtran}

\bibliography{IEEEabrv}

\end{document}